%% file: main.tex
\documentclass{article}

\usepackage[preprint]{neurips_2024}

\usepackage[utf8]{inputenc} 
\usepackage[T1]{fontenc}    
\usepackage{hyperref}       
\usepackage{url}            
\usepackage{booktabs}       
\usepackage{amsfonts}       
\usepackage{nicefrac}       
\usepackage{microtype}      
\usepackage{xcolor}         

\usepackage{graphicx}
\usepackage{balance}  
\usepackage{amsmath}
\usepackage[linesnumbered,ruled,vlined]{algorithm2e}
\usepackage{algpseudocode}
\usepackage{latexsym}
\usepackage{multirow}
\usepackage{multicol}
\usepackage{color}
\usepackage{dashrule}
\usepackage{extarrows}
\usepackage{dsfont}
\usepackage{centernot}
\usepackage{hyperref}
\usepackage{natbib}
\usepackage{fancybox}
\usepackage[utf8]{inputenc} 
\usepackage[T1]{fontenc}    
\usepackage{hyperref}       
\usepackage{url}            
\usepackage{booktabs}       
\usepackage{amsfonts}       
\usepackage{amsmath}
\usepackage{nicefrac}       
\usepackage{microtype}      
\usepackage{xcolor} 
\usepackage{amsmath}
\usepackage{lipsum}
\usepackage{colortbl}
\usepackage{subcaption}
\usepackage{tabularx}
\usepackage{makecell}
\usepackage{wrapfig}
\usepackage{bm}

\usepackage{todonotes}
\usepackage{multirow}
\usepackage{xcolor} 
\usepackage{bbm}
\usepackage{mdframed}
\definecolor{navyblue}{RGB}{0,0,128} 
\definecolor{vennred}{RGB}{255,0,0} 
\usepackage{xcolor}

\usepackage{xcolor}

\definecolor{nav}{RGB}{0,0,128}

\title{A Controlled Study on Long Context  Extension \\ and Generalization in LLMs
}

\author{Yi Lu$^2$\thanks{Equal contribution. $^+$ Correspondence author}~ \quad Jing Nathan Yan$^{1*}$ \quad Songlin Yang$^3$ \quad Justin T. Chiu$^1$ \\ \textbf{Siyu Ren$^2$ \quad Fei Yuan$^2$ \quad Wenting Zhao$^1$  \quad Zhiyong Wu$^{2+}$ \quad Alexander M. Rush$^1$}\\
$^1$Cornell University, $^2$Shanghai AI Lab, $^3$Massachusetts Institute of Technology
}

\begin{document}
\maketitle
\begin{abstract}

Broad textual understanding and in-context learning require language models that utilize full document contexts. Due to the implementation challenges associated with directly training long-context models, 
many methods have been proposed for extending models to handle long contexts. However, owing to differences in data and model classes, 
it has been challenging to compare these approaches, leading to uncertainty as to how to evaluate long-context performance and whether it differs from standard evaluation. We implement a controlled protocol for extension methods with a standardized evaluation, utilizing consistent base models and extension data. Our study yields several insights into long-context behavior. First, we reaffirm the critical role of perplexity as a general-purpose performance indicator even in longer-context tasks. Second, we find that current \textit{approximate} attention methods systematically underperform across long-context tasks. Finally, we confirm that exact fine-tuning based methods are generally effective within their extension range, whereas extrapolation remains challenging. All codebases, models, and checkpoints are made available open-source via \url{https://github.com/Leooyii/LCEG}, promoting transparency and facilitating further research in this critical area of AI development. 
\end{abstract}

\input{sections/introduction}
\input{sections/related_work}
\input{sections/background}
\input{sections/methodology}

\input{sections/experiments}
\input{sections/analysis}

\input{sections/conclusion}
\bibliographystyle{unsrtnat}
\bibliography{references}

\clearpage
\input{sections/appendix}

\end{document}

%% file: sections/introduction.tex
\section{Introduction}

The pretraining data scale of large language models (LLMs) has expanded greatly in recent years with open models trained up to 15T tokens~\citep{llama3modelcard}. Implementation challenges make it difficult to fully train models with longer context windows during pretraining~\citep{liu2023ring}.
Still, long-context windows are considered central, as they enable LLMs to perform tasks that require more extensive textual understanding, such as utilizing information from textbooks~\citep{tanzer2024a}, summarizing novels~\citep{kryscinski2022booksum}, and engaging in many-shot learning~\citep{bertsch2024context,li2023context}.

As a trade-off, researchers have proposed \textit{context extension}, where an LLM initially pretrained on standard sequences is adapted for significantly longer context lengths~\citep{chen2023extending, peng2023yarn, han2023lminfinite, bloc97}. These methods differ in the type of attention used and in post-training adaptation techniques. 
They vary in complexity, training requirements, and qualitatively exhibit significantly different performance profiles.

Unfortunately, there is a relatively poor understanding of the quantitative rankings of these different methodologies. Owing to the perceived challenges of evaluation, several new metrics, such as long context perplexity~\citep{chen2023extending, chen2023longlora, han2023lminfinite}, and retrieval accuracy \citep{mohtashami2023landmark, NeedleInAHaystack} have been introduced~\citep{bai2023longbench, an2023leval}. However, the differences in long-context extension procedures make it hard to calibrate these metrics while controlling for other factors. 

In this work, we implement a controlled protocol for context extension. The aim is to compare context extension while removing spurious factors that impact LLM ability.

\textbf{Modeling}: We standardize on the same base model for all experiments. Different base models behave significantly differently,  making it challenging to draw general conclusions. For instance, past work evaluates LM-Infinite~\citep{han2023lminfinite} on LongBench~\citep{bai2023longbench} using different base models \citep{xiao2024infllm,lu2024longheads}. 

\textbf{Extensions}: We implement a range of context extension methods within the same framework. We use a standardized recipe to eliminate potential gains from tailored hyperparameters. We additionally fix the post-training data for each method, utilizing an identical and open-sourced training corpus~\citep{fu2024data, chen2023longlora}.

\textbf{Metrics}: We look at both intrinsic metrics, such as perplexity, and extrinsic properties, such as downstream task performance~\citep{hsieh2024ruler, NeedleInAHaystack, bai2023longbench}. We consider metrics within the extension length as well as an extrapolation to longer contexts.

Our study identifies several takeaways for future research. First, contrary to some of the arguments for needing new metrics, our findings indicate a strong correlation between perplexity and downstream task performance for exact fine-tuned methods in controlled studies. While some approximate attention methods are unbalanced between perplexity and downstream task performance, there is a strong correlation between the two for most extension tasks. 

Second, we find relatively poor results for approximate attention methods. While they can handle longer length contexts, there generally is a trade-off in terms of accuracy for most of our benchmarks. Exact frozen methods also tend to degrade model performance, showing high sensitivity to hyperparameters and often failing with a general training recipe. 

Finally, continual fine-tuning with exact attention generally works well, particularly within the extended context length. Specifically, Dynamic NTK~\citep{emozillareddit} works best among these methods. Extrapolation to longer lengths remains a challenging problem. 
Our training code, models, and checkpoints will be open-sourced to support further research in this area.

%% file: sections/related_work.tex
\section{Related Work}
\paragraph{Long Context Methods}
We divide extension methods into three broad classes: exact attention, approximate attention, and context compression.
Exact attention methods augment the parameterization of attention. Position interpolation (PI)~\citep{chen2023extending}, NTK-aware~\citep{bloc97}, Dynamic NTK~\citep{emozillareddit}, YaRN~\citep{peng2023yarn}, and CLEX~\citep{chen2024clexcontinuouslengthextrapolation}, all based on RoPE \citep{Su_Lu_Pan_Wen_Liu_2021}, design position embeddings for length extension.
These methods may be applied with fine-tuning or to frozen models.
Other exact attention methods focus on training-time improvements, such as contrastive training \citep{tworkowski2023focused} .
Approximate attention methods introduce structured attention approximations that minimize the computational cost of length extension.
\cite{chen2023longlora} introduced the use of LoRA~\citep{DBLP:journals/corr/abs-2106-09685} and a specialized local attention mechanism to reduce further the computational overhead of further fine-tuning with long context.
Other approaches break the text into chunks and utilize a well-designed "chunk representation" to retrieve relevant chunks for attention~\citep{mohtashami2023landmark, xiao2024infllm, lu2024longheads}.
LM-Infinite and StreamLLM \citep{han2023lminfinite,xiao2023efficient} retain only a few tokens from the beginning of the text and a local window to keep the attention window within the pretrained length.
\citet{xu2024retrieval} focuses on using retrievers to retrieve relevant blocks from long documents.
Finally, context compression methods, which we do not explore in this work, reduce length extension to length compression via a summarization step
\citep{jiang2023longllmlingua,li2023compressing}.
\paragraph{Long Context Evaluation Benchmarks}
The Long Range Arena (LRA)~\citep{tay2020long} is an early efforts in evaluating the proficiency of processing long contexts in different modalities. Since then, a growing number of benchmarks have emerged, including LongBench~\citep{bai2023longbench}, LEval \citep{an2023leval}, and LooGLE \citep{li2023loogle}. These benchmarks are a mixture of diverse downstream tasks explicitly tailored to assess the capabilities of LLMs in understanding and generating lengthy contexts. Among these benchmarks, LongBench stands out for its inclusion of diverse sequences with varying lengths, distributions, patterns, languages, and domains, enabling a comprehensive, nuanced evaluation. In addition to evaluating LLMs' performance on downstream NLP tasks, there is another line of benchmarks that specifically focuses on assessing particular aspects of long context processing ability~\citep{liu2023lost,hsieh2024ruler}. For instance,  \citet{mohtashami2023landmark} propose the passkey retrieval task to challenge a language model to accurately locate and retrieve a simple passkey (a five-digit random number) in a long context sequence. Similarly, the Needle in a Haystack \citep{NeedleInAHaystack} test requires the model to accurately recite the information from a specified sentence(the "needle") from a long document.
However, most existing works mainly focus on evaluating mainstream commercial models (e.g. GPT-4 and Claude), open-source base models, or just perform individual evaluations of a few long context methods. There is a lack of comprehensive, yet controlled evaluation on long-context extension techniques themselves. 

%% file: sections/background.tex
\section{Context Extension Methods}
\label{subsec:Long Context Methods}

\subsection{Background: Attention and RoPE}

The bottleneck in long context modeling in Transformers is attention. Attention is defined over 
$C$ embeddings $\mathbf{X} = [\mathbf{x}_1, \mathbf{x}_2, \ldots, \mathbf{x}_C]^\top \in \mathbb{R}^{C \times d}$ where $d$ is the model dimension. Learned weight matrices $\mathbf{W}_v \in \mathbb{R}^{d \times d_k}$, $\mathbf{W}_q \in \mathbb{R}^{d \times d_k}$, and $\mathbf{W}_k \in \mathbb{R}^{d \times d_k}$ are used to transform these inputs where $d_k$ is the projected hidden dimension.  The attention mechanism itself computes the attention matrix and applies it to produce a weighted sum of the value vectors:
\begin{equation}
\text{Attention}(\mathbf{Q}, \mathbf{K}, \mathbf{V}) = \mathbf{A} \mathbf{V} = \text{softmax}\left(\frac{\mathbf{Q} \mathbf{K}^\top}{\sqrt{d_k}}\right)\mathbf{V}.
\end{equation}
Basic attention was originally defined with: $\mathbf{Q} = \mathbf{X} \mathbf{W}_q, \mathbf{K} = \mathbf{X} \mathbf{W}_k, \mathbf{V} = \mathbf{X} \mathbf{W}_v
$. However, this approach does not directly encode the relative position of keys and values.

Rotary Position Embeddings (RoPE) \citep{su2024roformer} encode positional information by applying a phase rotation to each element of the embedding vectors. Formally, we define a transformation $\mathbf{f}$:

\begin{equation}
\mathbf{f}_\mathbf{W}(\mathbf{x}_i, \bm{\theta}) = \mathbf{R}(\bm{\theta}, i)\mathbf{W}^\top \mathbf{x}_i
\end{equation}

Here $\mathbf{x}_i \in \mathbb{R}^{d_k}$ is an embedding for position $i$, $\mathbf{W}$ is a projection matrix, and $\bm{\theta} \in \mathbb{R}^{d_k / 2}$ is a frequency basis. The function is defined based on the rotary position matrix: 

\begin{equation}
\mathbf{R}(\bm{\theta},i)= \begin{pmatrix}
\cos i\theta_1 & - \sin i\theta_1 &  \cdots & 0 & 0 \\
\sin i\theta_1 & \cos i\theta_1 & \cdots & 0 & 0 \\
\vdots \\ 
0 & 0 &  \cdots & \cos i\theta_\frac{d_k}{2}  & - \sin i\theta_\frac{d_k}{2}  \\
0 & 0 &  \cdots & \sin i\theta_\frac{d_k}{2}  & \cos i\theta_\frac{d_k}{2}  \\
\end{pmatrix}
\end{equation}

 


Due to the arrangement of frequencies, this matrix has the property that $\mathbf{R}(\bm{\theta},n - m) = \mathbf{R}(\bm{\theta},m)^\top\mathbf{R}(\bm{\theta},n)$ by Ptolemy's identity. We redefine the query-key product between two positions $m$ and $n$ as, 

\begin{align}\label{equ:rope}
\mathbf{q}^\top_m{\mathbf{k}_n}&=\mathbf{f}_{\mathbf{W}_q}(\mathbf{x}_m, \bm{\theta})^\top\mathbf{f}_{\mathbf{W}_k}(\mathbf{x}_n, \bm{\theta}) \\ 
& = \left(\mathbf{R}({\bm{\theta},m})\mathbf{W}^\top_q\mathbf{x}_m \right)^\top \left( \mathbf{R}(\bm{\theta},n)\mathbf{W}^\top_k\mathbf{x}_n \right) \\
&= \mathbf{x}_m^\top\mathbf{W}_q \mathbf{R}(\bm{\theta},n - m) \mathbf{W}_k^\top\mathbf{x}_n 
\end{align}

In this way, the relative positional information $n-m$
is implicitly injected into the query and key product, thus the attention score.

The standard RoPE transformation, $f_\mathbf{W}(x_i,\bm{\theta})$, sets the components
${\theta}_j = b^{-\frac{2j}{d_k}}$ with base $b=10000$.





\subsection{Adjusting the Frequency of RoPE for Long Context Extension}

We consider four methods for performing length extension on RoPE embeddings: Position Interpolation (PI)~\citep{chen2023extending}, NTK-RoPE~\citep{emozillareddit}, YaRN~\citep{peng2023yarn} and CLEX~\citep{chen2024clexcontinuouslengthextrapolation}. In this section our goal is to extend a method trained to extend position embeddings for context length $C$ to length $C' >> C$. The methods in this section perform this extension by scaling the frequencies with the \textit{base scaling vector} $\bm\alpha\in\mathbb{R}^{\frac{d_k}{2}}$:
\begin{equation}
\mathbf{f}_\mathbf{W}(x_i) = \mathbf{f}(x_i, \bm\alpha \odot \bm{\theta}).
\end{equation}

\textbf{Linear Position Interpolation (PI)} decreases the frequencies of the basis functions so that more tokens fit within each period. 
PI is implemented by setting the components of the base scaling vector to
\begin{equation}
\alpha_j^{\text{PI}} = \frac{C}{C'} = \frac{1}{t}.
\end{equation}
where $t=\frac{C'}{C}$ is target length ratio. PI has been integrated into many open-source models such as LLaMA2-7B-32K~\citep{together-instruct}, Vicuna-7B-v1.5~\citep{vicuna2023}, and LongAlpaca~\citep{chen2023longlora}.

\textbf{Neural Tangent Kernel Interpolation RoPE (NTK-RoPE) } builds on linear position interpolation by introducing a per-dimension scaling factor. Inspired by findings from the NTK literature that show that high-frequency features are difficult for MLPs to learn, NTK-RoPE preserves high-frequency features while extending the period of low-frequency features. This is accomplished via a dimension-dependent base scaling vector $\bm\alpha$:
\begin{equation}
 \alpha_j^{\text{NTK-RoPE}} = \kappa^{-\frac{2j}{d_k}},
\end{equation}
where $\kappa = \left(t\right)^{\frac{d_k}{d_k-2}}$ so that the lowest frequency is scaled to match PI and the highest frequency remains the same as in RoPE.



An extension to this approach, Dynamic NTK-RoPE suggests that instead of fixing scaling based on a set ratio $s$ for all examples during inference, the formula should adapt to the current context length for a specific example. We followed the set up of \citet{fu2024data} for Dynamic NTK-RoPE. More details can be found in the Appendix \ref{appendix:imp_details}.

\textbf{YaRN}, another RoPE extension method, uses ``NTK-by-parts" interpolation strategies across different dimensions of the embedding space and introduces a temperature factor to adjust the attention distribution for long inputs. 

\begin{equation}
 \alpha_j^{\text{YaRN}} = (\left(1 - {\gamma}_j\right) \frac{1}{t} + {\gamma}_j) / \sqrt{T}
\end{equation}



We use a ramp vector $\bm{\gamma}$ to determine the interpolation between the $\frac{1}{t}$ and the original frequency base. The interpolation gating is set based on the frequency for the dimension $j$. 

\begin{equation}
\gamma_j = 
\begin{cases}
    0, & \text{if } {\theta}_j < p, \\
    1, & \text{if } {\theta}_j > q, \\
    \frac{{\theta}_j - p}{q - p}, & \text{otherwise}.
\end{cases}
\end{equation}

The values of $p, q, T$ can be tuned as needed. 

Other methods such as \textbf{CLEX}~\cite{chen2024clexcontinuouslengthextrapolation} models the scaling vectors as a dynamical system, intending to learn target-length dependent scaling vectors.

\subsection{Approximate Attention}
An alternative approach is to modify the attention function itself. Instead of exactly computing each longer attention, these methods select a subset of positions to attend to. We consider four well-known methods based on three different attention mechanisms: sparse attention, sliding window attention, and retrieval attention.

\textbf{LongLoRA}~\citep{chen2023longlora} avoids computing attention ranges over $C'$ by only computing the block-diagonl part of attention. Formally, given a sequence length of $C'$, LongLoRA divides it into $M$ blocks of size $B$, resulting in a sparse attention matrix $\mathbf{A} \in \mathbb{R}^{C'\times C'}$ with a block-diagonal structure:
\begin{equation}
    \mathbf{A} = \begin{bmatrix}
        \mathbf{A}_1 & \mathbf{0} & \cdots & \mathbf{0} \\
        \mathbf{0} & \mathbf{A}_2 & \cdots & \mathbf{0} \\
        \vdots & \vdots & \ddots & \vdots \\
        \mathbf{0} & \mathbf{0} & \cdots & \mathbf{A}_M
    \end{bmatrix}
\end{equation}

where $\mathbf{A}_i \in \mathbb{R}^{B \times B}$ is the attention matrix for the $i$-th block. In addition, they shift the blocks for half of the heads enabling the information flow between groups via shifting. Notably, while they employ local attention during the fine-tuning phase, full attention is still adopted during the inference stage.


\textbf{Landmark Attention}~\citep{mohtashami2023landmark} addresses the challenge of attending over long sequences by breaking the input sequence into chunks and using trainable ``landmark" tokens to summarize these chunks. The attention process is carried out in two stages. Given a sequence of $C'$ embeddings, divided into $M$ chunks, each of length $B$, the first step is to compute global attention between the query vectors $\mathbf{Q} \in \mathbb{R}^{C' \times d_k}$ (corresponding to all input embedding) and the landmark vectors $\mathbf{L} \in \mathbb{R}^{M \times d_k}$ (which represent the chunks). From this global attention, a set of $n$-most attended-to chunks is selected for further processing. Next, a local attention mechanism is applied within each of the selected chunks. For the $n$-th selected chunk, the key matrix for the chunk is denoted as $\mathbf{K}_n \in \mathbb{R}^{B \times d_k}$ and $\mathbf{Q}_n \in \mathbb{R}^{B \times d_k}$. The attention matrices are then computed as follows:
\begin{align}
\mathbf{A}^1 &= \text{softmax}\left(\frac{\mathbf{Q}\mathbf{L}^T}{\sqrt{d_k}}\right) \in \mathbb{R}^{C' \times M}, \\
\mathbf{A}^{2,n} &= \text{softmax}\left(\frac{\mathbf{Q}_n \mathbf{K}_n^T}{\sqrt{d_k}}\right) \in \mathbb{R}^{B \times B},
\end{align}
The final attention for each embedding is a combination of these two attention. This method efficiently scales attention mechanisms for long sequences by focusing on landmark tokens that summarize large parts of the sequence, followed by local attention within the relevant chunks.

\textbf{LM-Infinite}~\citep{han2023lminfinite} (a.k.a., Sliding Window Attention) maintains a sliding local window of size $M$ along with a fixed global memory of $G$ positions at the starting point of the given embedding. Given $C'$ embeddings,  attention is computed over the $M$ embeddings in its local window and $G$ embeddings in global memory. LM-Infinite replaces relative positional information $n-m$ with $\min(n-m, C)$ while computing the query and key product in Eq \ref{equ:rope}. Altogether, LM-Infinite reduces the complexity from $\mathcal{O}((C')^2)$ to $\mathcal{O}(C'(M+G))$ without the need to scale positional encoding.

\textbf{Self-Extend}~\citep{jin2024llm} maps the unseen positions in extended context length $C'$ to positions within the pretraining context length $C$ to avoid training. 
For each embeddings, Self-Extend chooses closest $M$ embeddings and  any embeddings beyond are divided into multiple groups. Each group contains $N$ embeddings. When performing query-key product between two positions $m$ and $n$ in Equation \ref{equ:rope}, the relative positional information $n-m$ is replaced by $r$ which is computed by scaling $n-m$ w.r.t $M$ and $N$:
\begin{equation}
r=
\begin{cases}
    n-m, & n-m \leq M, \\
    M + \left\lfloor\frac{n-m}{N}\right\rfloor - \left\lfloor\frac{M}{N}\right\rfloor, & n-m > M. \\
\end{cases}
\end{equation}
where $\lfloor \cdot \rfloor$ denotes the floor division. The maximum extended context length $C'$ is $(C-M) \cdot N + M$.



%% file: sections/methodology.tex
\section{Long-Context Extension Protocol}


\textbf{Base Model}
All models start from an identical LLaMA2-7B base checkpoint \citep{touvron2023llama}. This approach removes potential biases from the checkpoint basis. Note that to study the influence of base model, particularly the size of base model, we also use Phi-2-base checkpoint \citep{javaheripi2023phi} for context extension, to verify whether the trends and analyses we observed are consistent across different base models, thereby avoiding potential over-generalization. The results of Phi-2-base are reported in Appendix \ref{appendix:phi2_result}.

\textbf{Fine-Tuning}
We sample 1B tokens from a long-context data mixture from \citet{fu2024data} to achieve state-of-the-art context extension performance. The details of the data are reported in Appendix \ref{appendix:data_construct}. We focus on extending the context window from 4k to 32k, as most downstream tasks require contexts under 32k.

We maintain a fixed training recipe to ensure consistency across all models \citep{chen2023longlora}. We follow existing practices by keeping an exponential moving average (EMA) of model weights with a constant decay. Most training hyperparameters are based on \citep{fu2024data}, with the learning rate set to $2\times10^{-5}$. We use a linear learning rate warm-up and zero weight decay, utilizing 8 NVIDIA A100 GPUs.

For LongLora, we fine-tune the weights of the LoRA adapter with trainable embedding and normalization, then merge these trainable weights with the LLaMA2 base model for evaluation. For Landmark Attention, the training context length is set to 512, with a block size of 64. For CLEX, we set the max scale factor to 32 and use the SiLU activation function.

We reuse the original scale factor to maintain consistency for NTK, YaRN, and Position Interpolation methods. However, this base factor significantly degrades continual fine-tuned models, particularly causing performance deterioration in shorter sequences. Therefore, we conduct a grid search to determine a better scale factor for different input lengths for NTK-RoPE method. Based on our findings, we follow and improve upon \citet{fu2024data} to set the scale factor for NTK-RoPE method. The scale factor and its relationship with perplexity are reported in the Appendix \ref{appendix:ntk_scale_factor}. Please refer to the Appendix \ref{appendix:imp_details} for detailed hyperparameter setups.

\textbf{Metrics}
We consider two sets of intrinsic metrics. The first is based on \textit{perplexity}. 
We use the book corpus PG19 ~\citep{rae2019compressive} and the Proof-pile dataset~\citep{azerbayev2023proofnet} to evaluate the long sequence language modeling performances. Following \citet{press2022train}, all perplexity evaluations are calculated using a sliding window with a window size of 256. 

The second is based on \textit{retrieval}. We focus on the needle in the haystack task~\citep{NeedleInAHaystack}(NIAH).  NIAH involves identifying a specific, relevant piece of information (the "needle") within a large set of irrelevant data (the "haystack"). This task is commonly used to test the precision and recall capabilities of LLMs in scenarios where the relevant data is sparse and surrounded by a significant amount of noise. Additionally, we evaluate with RULER~\citep{hsieh2024ruler}. RULER enhances the standard NIAH test by incorporating variations with different types and quantities of needles. Additionally, it introduces new task categories, such as multi-hop tracing and aggregation, to evaluate behaviors beyond simple context-based searching. 

For extrinsic metrics, we consider a collection of tasks. LongBench~\citep{bai2023longbench} is a family of bilingual, multitask evaluations for long-context understanding widely used in measuring the long-context abilities of LLMs~\citep{jin2024llm,xiao2024infllm,lu2024longheads}.
LongBench includes single-document question answering, multi-document QA, summarization, few-shot learning, and code completion.
We follow \citet{bai2023longbench} to evaluate the models on 32k context window sizes by truncating the prompt from the middle when the task length exceeds a designated context window size. We also consider the ManyShots tasks, where the long-context model will be given several examples as prompts. We use the Trec News~\citep{kontonis2024active} dataset for this task.

%% file: sections/experiments.tex
\section{Experimental Results}
\label{subsec:Long Context Benchmarks}

\subsection{Result Overview}

\input{tables/attention_category}

\begin{table}[t]
    \centering    
    \caption{Perplexity results of different methods on PG 19 and Proof-file. NTK-32K and NTK-64K refer to NTK-Dynamic, which requires finetuning on longer text. 
    Len refers to the longest-length examples seen at training or fine-tuning. Ex refers to the exact attention. All results are produced by our experiments. }\vspace{1em}
    \begin{tabular}{llll|cccccc}
        \toprule
 \multicolumn{4}{c|}{Model Details}& \multicolumn{6}{c}{ Eval Length}\\ 
 &  Len& Ex & Methods& {2k}& {4k}& {8k}& {16k}& {32k}&{64k}\\ \midrule
          \multicolumn{9}{c}{\textbf{PG19}}\\\midrule

        \multirow{3}{*}{Frozen} 
        & 4k & \checkmark &LLaMA2 & \textbf{6.61}& \textbf{6.30}& - & - & - & - \\ 
        & 4k & &LM-Infinite & \textbf{6.61}& \textbf{6.30}& 6.25 & 6.45 & 6.71 & 8.49 \\ 
        & 4k & \checkmark &NTK-Frozen & \textbf{6.61}& \textbf{6.30}& 6.82 & 7.94 & 14.52 & - \\
        & 4k & &Self-Extend & \textbf{6.61}	& 6.32 & 6.15 & 6.07 & 6.11 & 7.15 \\
        \midrule
        \multirow{6}{*}{Finetuned}  & 32k &\checkmark &PI & 6.88 & 6.52 & 6.27 & 6.08 & 5.95 & - \\ 
        & 32k & $\checkmark$ &NTK-32K& 6.63 & 6.32 & \textbf{6.09} & \textbf{5.92}& \textbf{5.79}& \textbf{5.76} \\ 
        & 32k & $\checkmark$ &YaRN & 6.70 & 6.39 & 6.16 & 6.01 & 5.93 & - \\
        & 32k & $\checkmark$ &CLEX & 6.85 &	6.62 &	6.14 & 5.93 & 5.82 & 5.79 \\
        & 32k & &LongLora & 12.80 & 11.52 & 10.70 & 10.18 & 9.89 & - \\ 
        & 32k & &Landmark & {8.15} & {8.14} & {8.14} & {8.11} & {8.13} & {8.15} \\ 
        \midrule
        & 64k & \checkmark &NTK-64K& 6.83 & 6.49 & 6.25 & 6.07 & 5.93 & 5.85 \\ \midrule
        \multicolumn{9}{c}{\textbf{Proof-file}}\\ \midrule
        \multirow{3}{*}{Frozen} 
        & 4k &$\checkmark$ &LLaMA2  & 3.34 & 3.04 & - & - & - & - \\ 
        & 4k & &LM-Infinite & 3.34 & 3.04 & 2.94 & 3.02 & 3.11 & 3.12 \\ 
        & 4k & \checkmark &NTK-Frozen & 3.34 & 3.04 & 2.91 & 3.09 & 4.06 & 12.65 \\ 
        & 4k & &Self-Extend & 3.35 &	3.06 &	2.88 &	2.78 &	2.75 &	2.90 \\
        \midrule
        \multirow{6}{*}{Finetuned}& 32k & \checkmark & PI & 3.34 & 3.03 & 2.83 & 2.68 & 2.58 & - \\ 
        & 32k & \checkmark& NTK-32K& \textbf{3.27} & \textbf{2.98} & \textbf{2.78} & \textbf{2.64} & \textbf{2.54} & \textbf{2.48} \\ 
        & 32k & \checkmark & YaRN & 3.29 & 3.00 & 2.81 & 2.68 & 2.59 & 106.38 \\ 
        & 32k & $\checkmark$ &CLEX & 3.37 &	3.10 &	2.80 &	2.65 &	2.55 &	\textbf{2.48} \\
        & 32k & & LongLora & 5.97 & 5.10 & 4.58 & 4.27 & 4.13 & - \\ 
        & 32k & & Landmark & {4.51} & {4.50} & {4.48} & {4.49} & {4.49} & {4.49} \\ 
        \midrule
        & 64k & \checkmark& NTK-64K& 3.33 & 3.03 & 2.83 & 2.69 & 2.58 & 2.51 \\ \bottomrule
    \end{tabular}%
    \label{tab:ppl:results}
\end{table}
Table \ref{tab:attention_category} overviews the results across both types of evaluation. The main result demonstrate that fine-tuned exact attention methods for long contexts, such as NTK-32K and YARN, consistently outperform approximate attention methods and frozen methods by a significant margin. This suggests that trading accuracy for speed in approximate attention methods can result in a loss of important reasoning capabilities, particularly for retrieval-based tasks. The performance disparity highlights the importance of exact attention mechanisms in maintaining high accuracy over extended contexts, emphasizing the need for careful consideration of attention mechanisms in model design for long-context tasks. We now consider each type of result in more detail.

\subsection{Intrinsic tasks}
\paragraph{Perplexity}
Table \ref{tab:ppl:results} shows perplexity scores across length. We see that continuous fine-tuning methods like PI, YaRN, and LongLora effectively keep low perplexity scores within the pre-training context length.  However, when the context length exceeds perplexity scores escalate once the context surpasses the pre-trained window. Only NTK and CLEX can generalize to unseen sequence length in both pretraining and continual finetuning.  Additionally, we find that exact attention maintains better perplexity than LoRA, which may reduce LongLora's ability. We also note that results on both PG19 and Proof-file gave nearly consistent conclusions.

\input{figures/needle_fig}

\begin{table*}[h]
\centering
\caption{RULER evaluation. Performance of models evaluated at length from 4k to 64k. Each score is computed by averaging the accuracy of 13 tasks. Train Len refers to the longest-length examples seen at continuous finetuning. }
\begin{tabular}{ll|c|ccccc}
\toprule
 & \bf Models & \begin{tabular}{c}\bf  Train\\\bf  Len\end{tabular} & \bf4k & \bf8k & \bf16k & \bf32k & \bf64k \\
\midrule
\multirow{3}{*}{Frozen} & LLaMA2 & 4k & 80.94 & - & - & - & - \\
& LM-Infinite & 4k & 81.05 & 30.01 & 18.02 & 12.34 & 10.56  \\
& NTK-Frozen & 4k & 81.14 & 44.45 & 14.79 & 0.72  & 0.91 \\
& Self-Extend & 4k & 65.03 & 50.73 & 44.02 & 29.50 & 9.34  \\
\midrule
\multirow{6}{*}{Finetuned} & PI & 32k & 84.56 & 76.04 & 69.64 & 57.66 & 0.00 \\
& NTK-32K & 32k & 86.58 & \textbf{77.75} & \textbf{70.01} & 59.42 & 46.26 \\
& YaRN & 32k & 79.12 & 65.60 & 54.21 & 36.95 & 0.00 \\
& CLEX & 32k & 50.18 & 63.93 & 64.35 & 52.17 & 30.61 \\
& LongLora & 32k & 10.58 & 6.37 & 3.67 & 3.53 & 0.00 \\
& Landmark & 32k & 22.37 & 17.52 & 16.31 & 13.56 & 14.15 \\
\midrule
& NTK-64K & 64k & \textbf{86.60} & 76.34 & 69.56 & \textbf{60.03} & \textbf{49.31}  \\
\bottomrule
\end{tabular}
\label{tab:ruler}
\end{table*}

\paragraph{Needle-in-the-haystack} NIAH results are shown in Figure \ref{fig:needle}.
Continuous finetuning approaches such as NTK, PI, and YaRN have successfully retrieved the "needle" within the pretraining length. Yet, only the NTK and CLEX method can retrieve the needle beyond the pretraining length, aligning with the perplexity results. The performance of the Exact Attention Method generally surpasses that of the Approximate Attention Methods. LM-Infinite and Landmark Excel are only within the local window, and they struggle to retrieve the intermediate text accurately. Regarding the Dynamic NTK method, NTK-F exhibits weak generalization when not trained. When trained on the same amount of data(1B), NTK-32K outperforms NTK-64K. When trained on 2B tokens, NTK-64K demonstrated a significant performance improvement, details are in Appendix \ref{appendix:needle}.

\input{tables/longbench}

\paragraph{RULER}
We test all models on 13 diverse tasks from the four RULER categories~\citep{hsieh2024ruler}. Each model is evaluated with 500 examples for lengths of 4k, 8k, 16k, 32k, and 64k. Results are compared with the Llama2-7B baseline in Table \ref{tab:ruler}. We observed a similar trend as in the NIHK task, where NTK family methods performed the best. NTK-32k maintained relatively good performance compared to other methods finetuned with a length cap of 32k. Performance of models on different length and breakdown by 13 subtasks can be found in Appendix \ref{appendix:ruler_result}.

\subsection{Extrinsic tasks}

\paragraph{LongBench}
The evaluation results of most methods on LongBench are presented in Table \ref{tab:longbench_result}, and results on all methods are presented in Appendix \ref{appendix:longbench}. Both LM-Infinite and Landmark Attention exhibit significant performance degradation compared to the base model. In contrast, the NTK, PI, and YaRN methods have successfully maintained their performance at 32k, demonstrating comparable results among these methods. This suggests that PI and YaRN perform similarly in downstream tasks, while the NTK family of models remains stable.

Notably, the LongLoRA method, which utilizes LoRA, also experiences a performance decline relative to the base checkpoint, LLaMA2. We argue that this may be due to the sensitivity of the training procedures for LongLoRA, and we acknowledge this in our limitation discussion section.

Furthermore, the overall performance on LongBench has not shown significant improvement over LLaMA2.We hypothesize that this is due to the average length of LongBench test data (approximately 7.5k) being considerably shorter than the 32k context window of the long-context methods.

\begin{wrapfigure}{r}{0.5\textwidth}
    \centering\includegraphics[width=.5\textwidth]{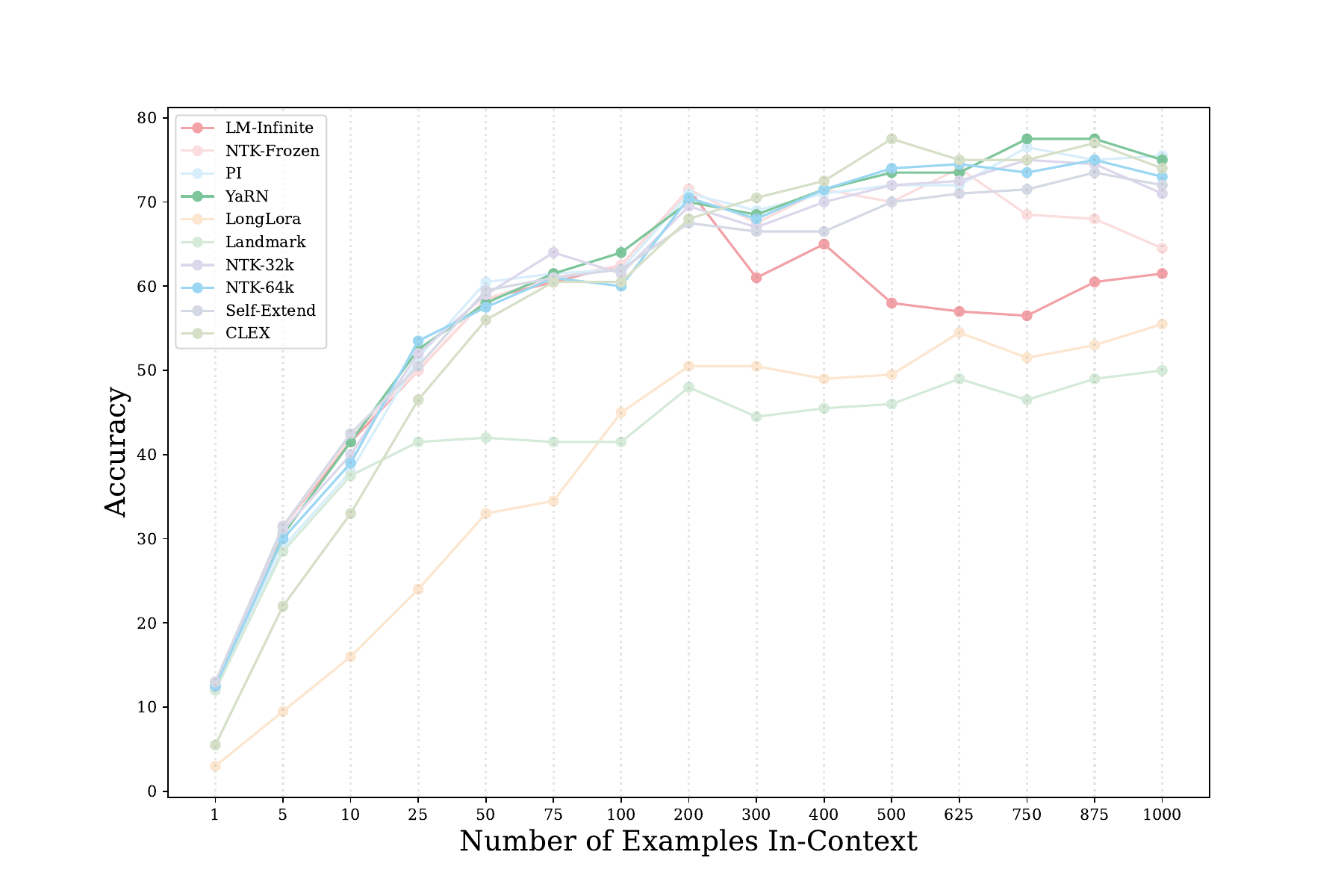}
    \caption{Many-shot in-context learning accuracy on TREC News.}
    \label{fig:manyshots}
    \vspace{-4em}
\end{wrapfigure}

\paragraph{Many-shot In-Context Learning with Trec News}
We evaluate TREC News~\citep{Li_Roth_2002} with 1 to 1000 in-context examples. Performance improves with more examples in Figure \ref{fig:manyshots}. Exact Attention methods show significant gains from 10 to 50 examples (+44.0\%) and 100 to 1000 examples (+25.9\%), with slower growth from 50 to 100 examples (+5.7\%). Approximate Attention methods consistently underperform. Performance gains align with model perplexity in longer contexts; NTK-Frozen excels with fewer examples but underperforms with more.

%% file: tables/attention_category.tex



\begin{table*}[t]
\centering
\caption{Overview of results across different extension types.}
\begin{tabular}{cc | c | c c c c c}
\toprule
\multicolumn{2}{c|}{\textbf{Attention} \textbf{Mechanisms}}& \textbf{Model} & \textbf{PPL} & \textbf{Needle} & \textbf{Mshots} & \textbf{LongB} & \textbf{RULER}\\
\midrule
\multirow{5}{*}{\makecell[c]{Exact\\ Attention}} & Frozen & NTK-F & 14.52 & 18.8 & 64.5 & 25.54 & \phantom{0}0.72\\
\cmidrule(l){2-8}
& \multirow{4}{*}{\makecell[c]{Fine-Tuned}}& PI & 5.85 & 42.1 & \textbf{75.5} & 33.48 & 57.66 \\
& & YaRN & 5.85 & 46.7 & 75.0 & 33.45 & 36.95 \\
& & CLEX & 5.82 & 71.1 & 74.0     & 33.48 & 52.17 \\
& & NTK-32K & \textbf{5.79} & \textbf{83.7} & 71.0 & \textbf{35.32} & 59.42\\
& & NTK-64K & 5.93 & 69.1 & 73.0 & 34.30 & \textbf{60.03} \\
\midrule
\multirow{3}{*}{\makecell[c]{Approxi. \\Attention}}& Frozen& LM-Infinite & 6.71 & 23.9 & 61.5 & 25.84 & 12.34 \\
& & Self-Extend & 6.11 & 25.8 & 72.0 & 33.62 & 29.50 \\
\cmidrule(l){2-8}
& Fine-tuned & LongLora & 9.89 & 20.3 & 55.5 & 23.30 & \phantom{0}3.53\\
& & Landmark & 8.13 & 50.9 & 50.0 & 28.19 & 13.56 \\

\bottomrule
\end{tabular}
\label{tab:attention_category}
\end{table*}


%% file: figures/needle_fig.tex
\begin{figure*}[t!]
  \centering
  \includegraphics[width=1\textwidth]{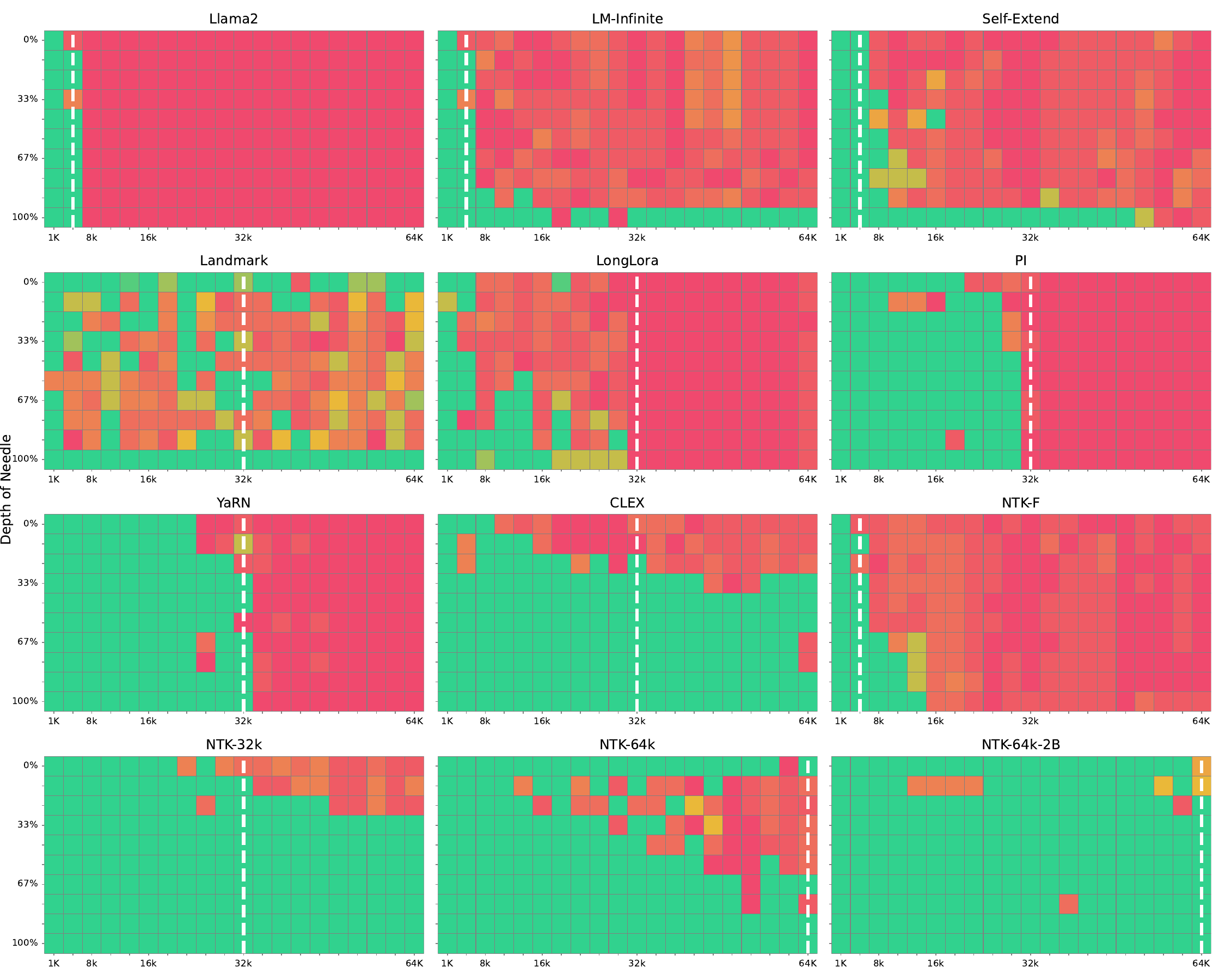}

  \caption{Needle in a Haystack evaluation. Green squares indicates a high retrieval success rate, the white dashed line denotes the longest length examples seen at training or finetuning, and the Y-axis represents the distance to the retrieved target.}\vspace{-1em}
  \label{fig:needle}
\end{figure*}

%% file: tables/longbench.tex
\begin{table*}[h]
    \centering
    \caption{LongBench results. N-32 and N-64 refer to NTK finetuned on 32K and 64K context lengths respectively. SE refers to Self-Extend. YN refers to YaRN. CX refers to CLEX. LLR refers to LongLora. Len refers to average length of the datasets. Train Len refers to the longest length examples seen at training or finetuning. Eval Len refers to the maximum length of the input prompt. \checkmark refers to whether the method is exact attention.}
    \begin{tabular}{lr|rrr|rrrrrrr}
    \toprule
 & & \multicolumn{3}{c|}{Frozen}& \multicolumn{7}{c}{Finetuned}\\
 & Len& Base  & N-F & SE & PI & N-32 & YN & CX & LLR & Land & N-64  \\ 
 &  & \checkmark & \checkmark &  & \checkmark & \checkmark & \checkmark &  & & \checkmark\\
      \midrule
    \multicolumn{2}{c|}{Train Len} & 4k & 4k & 4K & 32k & 32k & 32k & 32k & 32k & 32k & 64k \\ 
    \multicolumn{2}{c|}{Eval Len} &  4k & 32k & 32K & 32k & 32k & 32k & 32k & 32k & 32k & 32k \\
    \midrule
    \multicolumn{2}{l|}{\makebox[1cm][l]{NQA}\hfill 18k} & 21.09 & 3.88 & 23.49 & 23.02 & 23.73 & 19.82 & 24.19 & 12.07 & 12.47 & \textbf{24.31} \\
    \multicolumn{2}{l|}{\makebox[1cm][l]{QAP}\hfill 4k}  & 26.94 & 26.79 & 28.75 & 25.85 & \textbf{27.50} & 26.98 & 23.36 & 20.15 & 19.06 & 24.97 \\
    \multicolumn{2}{l|}{\makebox[1cm][l]{MQA}\hfill 5k}  & 32.42 & 29.82 & 32.66 & 35.10 & 38.22 & 37.11 & 40.83 & 24.50 & 21.86 & \textbf{40.60} \\
    \multicolumn{2}{l|}{\makebox[1cm][l]{HQA}\hfill 9k}  & 31.23 & 32.10 & 37.63 & 36.98 & \textbf{41.56} & 38.60 & 35.59 & 27.41 & 33.66 & 41.47 \\
    \multicolumn{2}{l|}{\makebox[1cm][l]{WQA}\hfill 5k}  & 25.75 & 22.34 & 30.70 & 29.38 & \textbf{31.58} & 30.63 & 28.24 & 21.46 & 24.94 & 28.62 \\
    \multicolumn{2}{l|}{\makebox[1cm][l]{MSQ}\hfill 11k}  & 10.55 & 8.84 & 15.73 & 16.80 & 17.41 & \textbf{22.08} & 17.12 & 11.46 & 11.41 & 18.24 \\
    \multicolumn{2}{l|}{\makebox[1cm][l]{GR}\hfill 9k}  & 17.32 & 17.87 & 13.15 & 25.61 & \textbf{28.27} & 20.98 & 24.68 & 24.05 & 17.20 & 24.37 \\
    \multicolumn{2}{l|}{\makebox[1cm][l]{QSM}\hfill 11k}  & 21.28 & 15.35 & 20.20 & 21.19 & 21.52 & 20.66 & 21.55 & 17.66 & 18.83 & \textbf{21.65} \\
    \multicolumn{2}{l|}{\makebox[1cm][l]{MWS}\hfill 2k}  & 3.44 & 9.30 & 1.50 & 10.55 & 22.13 & 8.91 & 16.96 & 21.19 & 19.43 & \textbf{25.02} \\
    \multicolumn{2}{l|}{\makebox[1cm][l]{TRE}\hfill 5k}  & 66.00 & 67.50 & 69.00 & \textbf{71.00} & 69.00 & 69.00 & 67.50 & 50.00 & 49.00 & 69.00 \\
    \multicolumn{2}{l|}{\makebox[1cm][l]{TQA}\hfill 8k}  & 87.89 & 18.69 & 88.44 & 88.55 & 88.86 & \textbf{89.63} & 89.36 & 12.28 & 74.75 & 88.65 \\
    \multicolumn{2}{l|}{\makebox[1cm][l]{SMS}\hfill 6k}  & 41.70 & 32.46 & 43.76 & \textbf{43.35} & 42.21 & 44.25 & 43.02 & 13.45 & 40.38 & 41.59 \\
    \multicolumn{2}{l|}{\makebox[1cm][l]{PSC}\hfill 11k}  & 2.10 & 2.67 & 0.00 & 1.50 & 2.68 & 1.05 & 2.50 & \textbf{4.57} & 0.64 & 2.09 \\
    \multicolumn{2}{l|}{\makebox[1cm][l]{PSR}\hfill 9k}  & \textbf{9.00} & 3.77 & 4.50 & 4.50 & 4.62 & 3.79 & 8.50 & 3.50 & 2.50 & 6.50 \\
    \multicolumn{2}{l|}{\makebox[1cm][l]{LCC}\hfill 1k}  & \textbf{68.22} & 63.64 & 68.47 & 55.05 & 56.78 & 54.06 & 49.45 & 57.12 & 56.70 & 52.04 \\
    \multicolumn{2}{l|}{\makebox[1cm][l]{REP}\hfill 4k}  & \textbf{61.73} & 53.69 & 59.99 & 47.26 & 49.09 & 47.60 & 42.84 & 51.92 & 48.23 & 39.68 \\
    \cmidrule{1-12}
    \multicolumn{2}{l|}{\makebox[1cm][l]{Avg.}\hfill 7k} & 32.92 & 25.54 & 33.62 & 33.48 & \textbf{35.32} & 33.45 & 33.48 & 23.30 & 28.19 & 34.30 \\
    \bottomrule
    \end{tabular}
    \label{tab:longbench_result}
\end{table*}

%% file: sections/analysis.tex
\section{Analysis}



\begin{figure*}[t]
  \centering
  \includegraphics[width=1\textwidth]{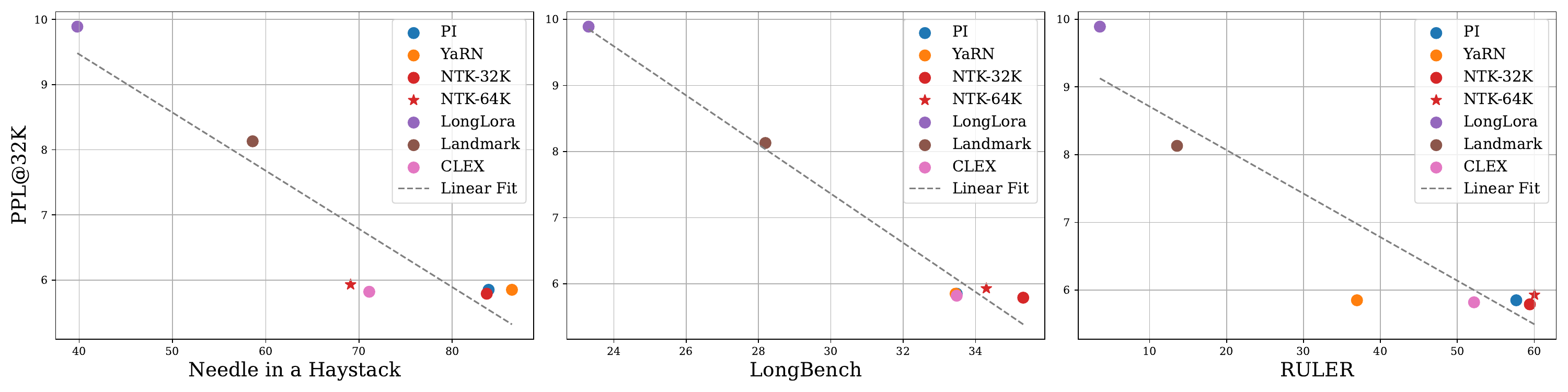}
  \caption{Perplexity and averaged downstream task accuracy for Needle in a haystack, LongBench and RULER.}\vspace{-1em}
  \label{fig:all:results}
\end{figure*}

\paragraph{Perplexity and Downstream Tasks} Several existing studies~\citep{sun2021longrange, an2023leval} suggest that perplexity may not consistently correlate with performance on long-range tasks. In Figure \ref{fig:all:results}, we plot the perplexity of models from our evaluated benchmarks against their performance. The figure shows a general correlation between perplexity and model performance across various tasks for exact attention methods. However, approximate attention methods, LongLora, and Landmark on RULER, exhibit slight deviations from this trend but still roughly fit into the linear relationship. We hypothesize that previous studies mentioned this disconnection due to the lack of controlled studies and the presence of noisy data.

\begin{wrapfigure}{r}{0.5\textwidth}
    \centering
    \includegraphics[width=.5\textwidth]{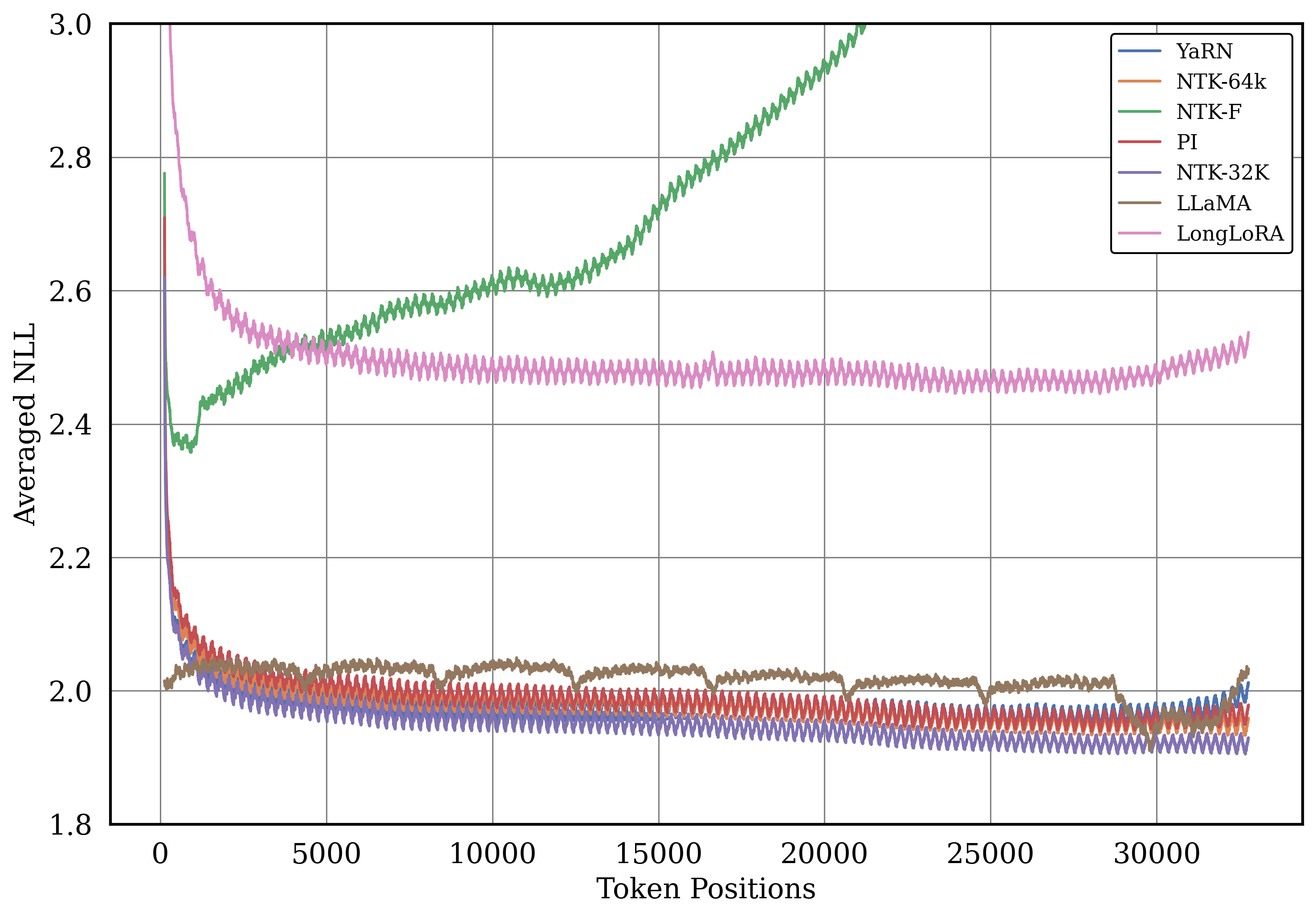}
    \caption{Averaged negative log-likelihood of different models broken down by context position.}
    \vspace{-2em}
    \label{fig:ppl:position}
\end{wrapfigure}

As depicted in Figure \ref{fig:needle}, although LM-Infinite achieves a good perplexity score at 32k, it fails to generalize beyond 4k. This limitation arises because the LM-Infinite method focuses on a context window of only 4k, resulting in poor retrieval ability for longer contexts. This suggests that we should consider downstream tasks at different lengths when evaluating perplexity for approximate attention methods.

\paragraph{Context extension hurts in the short term and gains in the long term} While context extension seems to improve perplexity, in Table \ref{tab:longbench_result}, we do not notice a significant gain in performance.

We hypothesize that while this dataset contains long tasks, the average length is much shorter than 32K. These methods seem to improve the ability to model language over the long term but hurt in the short term. To understand this better we compute the averaged negative likelihood of each position of YaRN, LLaMa2, and NTK-32K per position (with LLaMa2 seeing just tokens every 4k chunks) in Figure \ref{fig:ppl:position}.

\paragraph{NTK Generalizes Beyond 32k} In Figure \ref{fig:needle}, we observe that NTK-32K successfully generalizes to unseen sequence lengths beyond 32k in both Needle-in-the-Haystack and RULER tasks, performing on par with NTK-64K. In contrast, NTK-F demonstrates generalization up to 8k but fails to extend further. This suggests that while NTK methods may possess the capability to generalize to longer unseen sequences, their effectiveness is contingent upon conditions such as continual fine-tuning.

We find that up until 4K they all improve as expected with LLaMa2 having the best NLL. After 4K they all fluctuate in average, but we see a clear separation with Yarn and NTK taking into account the long context. At extremely long context NTK remains a strong model whereas Yarn becomes reverts to a similar performance as LLaMa2.

%% file: sections/conclusion.tex
\section{Limitations and Broader Impacts}
\noindent \textbf{Limitations} As we are limited by computing budget, we only experiment with Llama-2-7B as our base model. The findings in this paper may not generalize to other, possibly larger, base models. We only fine-tuned models to context sizes of 32k, and generalization behaviors to longer contexts may differ when training contexts are longer. We also acknowledge that the standardized training recipe with fixed hyperparameters may bias some models more than the other models.

\noindent \textbf{Broader Impacts}
This paper provides an in-depth evaluation of various methods for extending the context lengths of LLMs, offering insights into their performance in long-range settings. By standardizing evaluations with a consistent base model and extension data, this work provides clearer benchmarks for future work. We also open-source code and models to promote transparency. Our models share similar risks as standard LLMs, where they may be used to generate harmful content and misinformation.

\section{Conclusion}
In this paper, we used a standardized approach to assess the performance of various long-context methods in LLMs. 
Our study underscores the role of perplexity as a crucial, performance indicator at length and highlights the trade-offs inherent in different attention mechanisms. 
We shed light on the strengths and weaknesses of various approaches, providing valuable insights for future research. 
As part of our commitment to open science, all our resources, including codebases, models, and checkpoints, will be open-sourced upon acceptance, fostering future advancements in this pivotal area of AI research.

\section{Acknowledgements}

We thank Yao Fu, Yue Yu, Tianyu Gao,  Celine Lee, Woojeong Kim, Jack Morris, Junxiong Wang, and Oscar Yin for their suggestions and feedback. This work was supported by NSF CAREER \#2037519 and NSF \#1901030.

%% file: sections/appendix.tex
\section*{Appendix}
\subsection{Phi-2 for Context Extension}
\label{appendix:phi2_result}
We use another open-weight model, Phi-2-base~\cite{javaheripi2023phi} as the base point for context extension, to verify whether the trends and analyses we observed are consistent across different base models.
Using an identical training recipe, we re-train and re-evaluate seven models with Phi-2-base. 
\subsubsection{Perplexity on Proof-file of Phi-2-base}
We evaluate the perplexity score of Phi-2-base on seven models in Table \ref{tab:phi2_ppl}. Consistent with our observations in the original submission, continuous fine-tuning methods like PI, YaRN, and LongLora effectively maintain low perplexity scores within the pre-training context length. However, perplexity scores escalate once the context length exceeds the pre-trained window. Notably, only NTK could generalize to unseen sequence lengths during both pre-training and continual fine-tuning.

\begin{table}[h]
    \centering    
    \caption{Perplexity results of different methods on Proof-file with Phi-2-base. NTK-32K and NTK-64K refer to NTK-Dynamic, which requires finetuning on longer text. 
    Len refers to the longest-length examples seen at training or fine-tuning. Ex refers to the exact attention. All results are produced by our experiments. }\vspace{1em}
    \begin{tabular}{llll|cccccc}
        \toprule
 \multicolumn{4}{c|}{Model Details}& \multicolumn{6}{c}{ Eval Length}\\ 
 &  Len& Ex & Methods& {2k}& {4k}& {8k}& {16k}& {32k}&{64k}\\ \midrule
        \multirow{3}{*}{Frozen} 
        & 2k &$\checkmark$ &Phi-2-base  & \textbf{4.02} & 25.72 & 175.05 & - & - & - \\ 
        & 2k & \checkmark &NTK-Frozen & \textbf{4.02} & 3.73 & 4.07 & 5.49 & 12.58 & 36.68 \\ 
        & 2k & &Self-Extend & 4.08 & \textbf{3.70} & \textbf{3.48} & 3.42 & 3.48 & 3.73 \\ 
        \midrule
        \multirow{3}{*}{Finetuned}& 32k & \checkmark & PI & 7.53 & 6.75 & 6.25 & 5.97 & 5.83 & 45.00 \\ 
        & 32k & \checkmark& NTK-32K& 4.24 & 3.81 & 3.51 & \textbf{3.32} & \textbf{3.18} & \textbf{3.20} \\ 
        & 32k & $\checkmark$ &CLEX & 5.53 & 4.32 & 3.78 & 3.51 & 3.42 & 3.60 \\ 
        \midrule
        & 64k & \checkmark& NTK-64K& 4.63 & 4.14 & 3.82 & 3.61 & 3.47 & 3.38 \\ \bottomrule
    \end{tabular}%
    \label{tab:phi2_ppl}
\end{table}

\subsubsection{RULER of Phi-2-base}
We test all models on 12 diverse tasks (except QA-2) from the four Ruler \citet{hsieh2024ruler} categories in Table \ref{tab:phi2_ruler}. Consistently, NTK-32k maintains relatively strong performance compared to other methods fine-tuned with a length cap of 32k. The only exception was CLEX and NTK-32k at 64k, showing a slight drop in performance.

\begin{table}[h]
\centering
\caption{RULER evaluation on seven methods with Phi-2-base. Performance of models evaluated at length from 2k to 64k. Each score is computed by averaging the accuracy of 12 tasks. Train Len refers to the longest-length examples seen at continuous finetuning. }
\begin{tabular}{ll|c|cccccc}
\toprule
 & \bf Models & \begin{tabular}{c}\bf  Train\\\bf  Len\end{tabular} & \bf2k & \bf4k & \bf8k & \bf16k & \bf32k & \bf64k \\
\midrule
\multirow{3}{*}{Frozen} & Phi-2-base & 2k & 83.73 & - & - & - & - & - \\
& NTK-Frozen & 2k & \textbf{83.98} & 52.95 & 18.09 & 4.07 & 0.06 & 0.00 \\ 
& Self-Extend & 2k & 68.55 & 50.82 & 36.65 & 22.00 & 7.83 & 2.32 \\ 
\midrule
\multirow{3}{*}{Finetuned} & PI & 32k & 25.51 & 23.19 & 16.88 & 14.99 & 4.78 & 0.00 \\ 
& NTK-32K & 32k & 81.18 & 66.90 & 52.57 & \textbf{46.53} & \textbf{32.06} & 12.84 \\ 
& CLEX & 32k & 75.33 & \textbf{72.66} & \textbf{53.56} & 46.23 & 25.46 & 13.03 \\ 
\midrule
& NTK-64K & 64k & 78.73 & 59.87 & 47.56 & 41.87 & 25.66 & \textbf{17.69} \\ 
\bottomrule
\end{tabular}
\label{tab:phi2_ruler}
\end{table}

\subsection{Implementation Details}
\label{appendix:imp_details}
\paragraph{Training}~To maintain consistency across all models, we use a fixed training protocol \citep{chen2023longlora}. We adopt standard practices by applying an exponential moving average (EMA) to the model weights with a constant decay rate. The majority of our training hyperparameters are derived from \citep{chen2023longlora}, including a learning rate of $2\times10^{-5}$. We implement a linear warm-up for the learning rate and set the weight decay to zero, utilizing 8 NVIDIA A100 GPUs.

For LongLora, we fine-tune the LoRA adapter weights along with trainable embeddings and normalization, subsequently integrating these trained weights into the LLaMA2 base model for evaluation. For Landmark Attention, the training context length is 512, with a block size of 64. For CLEX, we set the max scale factor to 32 and use the SiLU activation function. 
We present the hyperparameter settings for different methods during the training stage in Table \ref{tab:train_hyper}.

\paragraph{Inference}~
We list the scale factors used for different length ranges during inference in Table \ref{tab:inference_hyper}. 
For NTK-RoPE, given the maximum observed length during training or inference, $C_{\text{test}}$, and the scaling hyperparameter $s$, we follow \citet{fu2024data} in replacing $t$ with $s \cdot \frac{\max(C', C_{\text{test}})}{C} - (s - 1)$, and set the hyperparameter $s$ to $\frac{C'}{2C}$ during both training and inference.
For LM-infinite, we set the global memory \( G = 10 \) and the local window \( M = 4096 \).
For Landmark Attention, the training context length is set to 512, with a block size of 64.
For Self-Extend, we set the local window size $M$ for neighbor tokens to 1024 and the group size $N$ to 64.

\begin{table}[h]
\centering
\caption{Hyperparameters for Different Long Sequence Methods in Training.}
\begin{tabular}{l|c|c|ccc}
\toprule
\bf Models & \begin{tabular}{c}\bf  Train\\\bf  Len\end{tabular} & \begin{tabular}{c}\bf  Train\\\bf  Tokens\end{tabular} & $\alpha$ & bsz & lr \\
\midrule
PI & 32k & 1B & 8.0 & 32 & 2e-5 \\
NTK-32K & 32k & 1B & 29.0 & 32 & 2e-5 \\
YaRN & 32k & 1B & 8.0 & 32 & 2e-5 \\
LongLora & 32k & 1B & 8.0 & 32 & 2e-5 \\
Landmark & 32k & 1B & - & 32 & 2e-5 \\
\midrule
NTK-64K & 64k & 1B & 57.0 & 32 & 2e-5  \\
\bottomrule
\end{tabular}
\label{tab:train_hyper}
\end{table}

\begin{table}[h]
\centering
\caption{Hyperparameters for the Scale Factor Different Long-context Methods in Inference.}
\begin{tabular}{l|ccccc}
\toprule
\bf Models & \bf4k & \bf8k & \bf16k & \bf32k & \bf64k \\
\midrule
Llama2 & - & - & - & - & - \\
LM-Infinite & - & - & - & - & -  \\
NTK-Frozen & 1.0 & 3.0 & 7.0 & 15.0 & 31.0 \\
\midrule
PI & 8.0 & 8.0 & 8.0 & 8.0 & 8.0 \\
NTK-32K & 29.0 & 29.0 & 29.0 & 29.0 & 61.0 \\
YaRN & 8.0 & 8.0 & 8.0 & 8.0 & 8.0 \\
LongLora & 8.0 & 8.0 & 8.0 & 8.0 & 8.0 \\
Landmark & - & - & - & - & - \\
\midrule
NTK-64K & 57.0 & 57.0 & 57.0 & 57.0 & 57.0  \\
\bottomrule
\end{tabular}
\label{tab:inference_hyper}
\end{table}

\subsection{Training Data Construction}
\label{appendix:data_construct}
We sample 1B tokens from a long-context data mixture following \cite{fu2024data}.
We use the SlimPajama~\citep{cerebras2023slimpajama} dataset for continuous finetuning. 
This dataset serves as an open-source replication of the LLaMA~\citep{touvron2023llama} pretraining data mixture. It comprises 82\% web data (sourced 67\% from CommonCrawl and 15\% from C4), 4.5\% code data (Github), 4.5\% Wikipedia content, 4.5\% books, 2.5\% Arxiv papers, and 2.0\% StackExchange content. 
We use per-source length-upsampling to sample 1B tokens from the datasets, which increases the portion of long sequences while keeping the domain mixture the same.
We packed all sampled data into chunks of the corresponding training length, regardless of document boundaries, following common practice\cite{touvron2023llama, fu2024data}.

\subsection{Longer Model Needs more Training Tokens}
\label{appendix:needle}
\begin{figure*}[t!]
  \centering
  \includegraphics[width=1\textwidth]{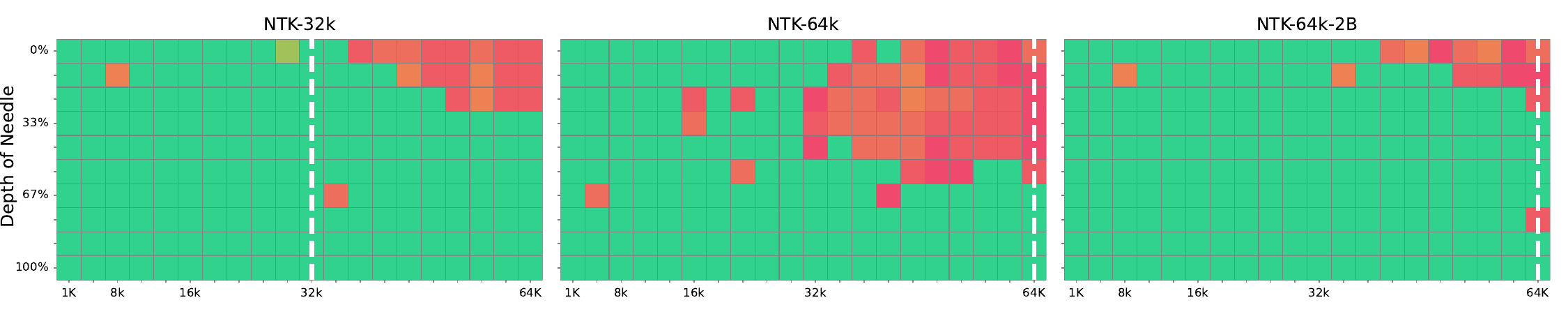}

  \caption{Needle in a Haystack evaluation. ``NTK-64-2B'' represents the NTK-64K model trained with 2B tokens. Green squares indicates a high retrieval success rate, the white dashed line denotes the longest length examples seen at training or finetuning, and the Y-axis represents the distance to the retrieved target.}\vspace{-1em}
  \label{fig:needle_more_tokens}
\end{figure*}

We observe that the performance of NTK-64K is not as good as NTK-32K. Consequently, we further sample 2B tokens from a long-context data mixture from \cite{fu2024data} for training and evaluate the model on the "Needle in A Haystack" task, as shown in Figure \ref{fig:needle_more_tokens}. Our NTK-64K model demonstrates a significant performance improvement when trained with more tokens, indicating that longer models require more tokens for effective training.

\subsection{RoPE Scale Factor for Dynamic NTK}
\label{appendix:ntk_scale_factor}
We observe that the scale factor significantly degrades NTK-Dynamic models, particularly causing performance deterioration in shorter sequences. Therefore, we conduct a grid search to determine a better scale factor for different input lengths.
The scale factor and its relationship with perplexity on PG19 are reported in Table \ref{tab:ntk_scale_factor}.

\begin{table}[h]
\centering
\caption{The scale factor and its relationship with perplexity on PG19. We only use the first 2 documents of PG19 to calculate the perplexity.}
\begin{tabular}{l|c|ccccc}
\toprule
\textbf{Models} & \multicolumn{1}{c|}{\textbf{Scale Factor}} & \textbf{4k} & \textbf{8k} & \textbf{16k} & \textbf{32k} & \textbf{64k} \\
\midrule
\multirow{6}{*}{NTK-Frozen} & 1 & \textbf{7.65} & 118.82 & NaN & NaN & NaN \\
& 3 & 8.19 & \textbf{7.99} & 57.15 & 386.02 & NaN \\
& 7 & 9.39 & 9.26 & \textbf{9.61} & 72.62 & 486.13 \\
& 15 & 11.53 & 12.04 & 12.98 & \textbf{20.15} & 180.59 \\
& 31 & 16.18 & 20.66 & 26.67 & 40.06 & \textbf{69.01} \\
& 63 & 30.22 & 48.78 & 69.89 & 89.75 & 118.59 \\
\midrule
\multirow{6}{*}{NTK-32K} & 1 & 12.64 & NaN & NaN & NaN & NaN \\
& 5 & 7.84 & 7.638 & 10.36 & NaN & NaN \\
& 13 & \textbf{7.686} & 7.459 & 7.25 & 8.35 & NaN \\
& 29 & 7.689 & \textbf{7.457} & \textbf{7.24} & \textbf{6.82} & 9.11 \\
& 61 & 7.8 & 7.565 & 7.34 & 6.91 & \textbf{6.63} \\
& 125 & 7.99 & 7.774 & 7.57 & 7.13 & 6.83 \\
\midrule
\multirow{5}{*}{NTK-64K} & 1 & 19.16 & NaN & NaN & NaN & NaN \\
& 9 & 8.02 & 7.79 & 7.63 & 22.6 & NaN \\
& 25 & \textbf{7.89} & \textbf{7.65} & \textbf{7.443} & 7.04 & 14.02 \\
& 57 & 7.922 & 7.67 & 7.44 & \textbf{7.01} & \textbf{6.75} \\
& 121 & 8.016 & 7.75 & 7.51 & 7.06 & 6.77 \\
\bottomrule
\end{tabular}
\label{tab:ntk_scale_factor}
\end{table}

\subsection{LongLora Validation}
To validate our LongLora \cite{chen2023longlora} implementation, we reproduced their Llama-2-7b-longlora-32k model following LongLora's training data and training recipe. We evaluated the perplexity for the corresponding length on PG19 and Proof-file in Table \ref{tab:longlora_validation}.
\begin{table}[h]
    \centering    
    \caption{Perplexity results of LongLora reported and our reproduction on PG 19 and Proof-file.}\vspace{1em}
    \begin{tabular}{l|ccccc}
        \toprule
        Method & {2k}& {4k}& {8k}& {16k}& {32k}\\ \midrule
        \multicolumn{6}{c}{\textbf{PG19}}\\\midrule
        Llama-2-7b-longlora-32k & 8.29 & 7.83 & 7.54 & 7.35 & 7.22 \\
        Our Reproduction & 8.10 & 7.69 & 7.43 & 7.28 & 7.32 \\ \midrule
        \multicolumn{6}{c}{\textbf{Proof-file}}\\ \midrule
        Llama-2-7b-longlora-32k & 3.35 & 3.01 & 2.78 & 2.61 & 2.50 \\
        Our Reproduction & 3.33 & 3.01 & 2.80 & 2.67 & 2.61 \\ \bottomrule
    \end{tabular}%
    \label{tab:longlora_validation}
\end{table}

\subsection{LongBench Results}
The evaluation results of all methods on LongBench are presented in Table \ref{tab:appendix_longbench_result}.
\label{appendix:longbench}
\begin{table*}[h]
    \centering
    \caption{LongBench results. N-32 and N-64 refer to NTK finetuned on 32K and 64K context lengths respectively. Inf refers to LM-Infinite. SE refers to Self-Extend. LLR refers to LongLora. AvgLen refers to average length of the datasets. Train Len refers to the longest length examples seen at training or finetuning. Eval Len refers to the maximum length of the input prompt. \checkmark refers to whether the method is exact attention.}
    \resizebox{\textwidth}{!}{ 
    \begin{tabular}{lr|rrrr|rrrrrrr}
    \toprule
 Exact& AvgLen& \multicolumn{4}{c|}{Frozen}& \multicolumn{7}{c}{Finetuned}\\
     & & Base  & Inf & N-F & SE & PI & N-32 & YaRN & CLEX & LLR & Land & N-64  \\ 
     &  & \checkmark &  & \checkmark &  & \checkmark & \checkmark & \checkmark &  & & \checkmark\\
      \midrule
    \multicolumn{2}{c|}{Train Len} & 4k & 4k & 4k & 4K & 32k & 32k & 32k & 32k & 32k & 32k & 64k \\ 
    \multicolumn{2}{c|}{Eval Len} &  4k & 32k & 32k & 32K & 32k & 32k & 32k & 32k & 32k & 32k & 32k \\
    \midrule
NQA     & \textbf{18,409}    & 21.09 & 10.39 & 3.88 & 23.49 & 23.02 & 23.73 & 19.82 & 24.19 & 12.07 & 12.47 & \textbf{24.31} \\
QAPR    & 3,619     & 26.94 & 22.58 & 26.79 & 28.75 & 25.85 & \textbf{27.50} & 26.98 & 23.36 & 20.15 & 19.06 & 24.97 \\
MFQA    & 4,559     & 32.42 & 26.19 & 29.82 & 32.66 & 35.10 & 38.22 & 37.11 & 40.83 & 24.50 & 21.86 & \textbf{40.60} \\
HPQA    & 9,151     & 31.23 & 16.13 & 32.10 & 37.63 & 36.98 & \textbf{41.56} & 38.60 & 35.59 & 27.41 & 33.66 & 41.47 \\
WMQA    & 4,887     & 25.75 & 20.64 & 22.34 & 30.70 & 29.38 & \textbf{31.58} & 30.63 & 28.24 & 21.46 & 24.94 & 28.62 \\
MSQ     & 11,214    & 10.55 & 5.26 & 8.84 & 15.73 & 16.80 & 17.41 & \textbf{22.08} & 17.12 & 11.46 & 11.41 & 18.24 \\
GR      & 8,734     & 17.32 & 13.43 & 17.87 & 13.15 & 25.61 & \textbf{28.27} & 20.98 & 24.68 & 24.05 & 17.20 & 24.37 \\
QMSM    & 10,614    & 21.28 & 6.10 & 15.35 & 20.20 & 21.19 & 21.52 & 20.66 & 21.55 & 17.66 & 18.83 & \textbf{21.65} \\
MNWS    & 2,113     & 3.44 & 3.63 & 9.30 & 1.50 & 10.55 & 22.13 & 8.91 & 16.96 & 21.19 & 19.43 & \textbf{25.02} \\
TREC    & 5,177     & 66.00 & 61.00 & 67.50 & 69.00 & \textbf{71.00} & 69.00 & 69.00 & 67.50 & 50.00 & 49.00 & 69.00 \\
TRVQA   & 8,209     & 87.89 & 81.40 & 18.69 & 88.44 & 88.55 & 88.86 & \textbf{89.63} & 89.36 & 12.28 & 74.75 & 88.65 \\
SMSM    & 6,258     & 41.70 & 15.07 & 32.46 & 43.76 & \textbf{43.35} & 42.21 & 44.25 & 43.02 & 13.45 & 40.38 & 41.59 \\
PSC     & 11,141    & 2.10 & 1.62 & 2.67 & 0.00 & 1.50 & 2.68 & 1.05 & 2.50 & \textbf{4.57} & 0.64 & 2.09 \\
PSR     & 9,289     & \textbf{9.00} & 4.00 & 3.77 & 4.50 & 4.50 & 4.62 & 3.79 & 8.50 & 3.50 & 2.50 & 6.50 \\
LCC     & 1,235     & \textbf{68.22} & 67.68 & 63.64 & 68.47 & 55.05 & 56.78 & 54.06 & 49.45 & 57.12 & 56.70 & 52.04 \\
REPO    & 4,206     & \textbf{61.73} & 58.27 & 53.69 & 59.99 & 47.26 & 49.09 & 47.60 & 42.84 & 51.92 & 48.23 & 39.68 \\
    \cmidrule{1-13}
    Average & 7,425 & 32.92 & 25.84 & 25.54 & 33.62 & 33.48 & \textbf{35.32} & 33.45 & 33.48 & 23.30 & 28.19 & 34.30  \\
    \bottomrule
    \end{tabular}
    }
    \label{tab:appendix_longbench_result}
\end{table*}

\subsection{RULER Subtasks Result}
\label{appendix:ruler_result}

The performance of models on different lengths and breakdowns by 13 subtasks are reported in Table \ref{tab:ruler_4k}(RULER on 4k),  Table \ref{tab:ruler_8k}(RULER on 8k), Table \ref{tab:ruler_16k}(RULER on 16k), Table \ref{tab:ruler_32k}(RULER on 32k) and Table \ref{tab:ruler_64k}(RULER on 64k).

\begin{table*}[h]
    \centering
    \caption{Ruler results on 4k context length. N-32 and N-64 refer to NTK finetuned on 32K and 64K context lengths respectively. Inf refers to LM-Infinite. SE refers to Self-Extend. LLR refers to LongLora. Train Len refers to the longest length examples seen at training or finetuning. Eval Len refers to the maximum length of the input prompt. \checkmark refers to whether the method is exact attention.}
    \resizebox{\textwidth}{!}{ 
    \begin{tabular}{l|rrrr|rrrrrrr}
    \toprule
    Exact& \multicolumn{4}{c|}{Frozen} & \multicolumn{7}{c}{Finetuned} \\
         & Base & Inf & N-F & SE & PI & N-32 & YaRN & CLEX & LLR & Land & N-64 \\
         & \checkmark & & \checkmark & & \checkmark & \checkmark & \checkmark & \checkmark & & & \checkmark \\
    \midrule
    Train Len & 4k & 4k & 4k & 4k & 32k & 32k & 32k & 32k & 32k & 32k & 64k \\
    Eval Len  & 4k & 4k & 4k & 4k & 4k  & 4k  & 4k  & 4k & 4k  & 4k  & 4k  \\
    \midrule
    NIAH\_S1 & 100.00 & 100.00 & 100.00 & 100.00 & 98.00 & 100.00 & 99.60 & 100.00 & 0.00  & 49.00 & 100.00 \\
    NIAH\_S2 & 100.00 & 100.00 & 100.00 & 100.00 & 99.80 & 100.00 & 88.60 & 100.00 & 0.00  & 20.60 & 100.00 \\
    NIAH\_S3 & 99.20  & 95.80  & 98.80  & 89.80  & 99.80 & 94.20  & 53.00 & 89.60  & 0.00  & 10.00 & 97.20  \\
    NIAH\_M1 & 99.20  & 98.80  & 99.20  & 79.00  & 99.20 & 99.20  & 62.60 & 95.80  & 0.00  & 10.60 & 98.00  \\
    NIAH\_M2 & 88.00  & 88.00  & 88.20  & 26.00  & 95.40 & 97.40  & 14.00 & 83.20  & 0.00  & 6.80  & 97.00  \\
    NIAH\_M3 & 61.40  & 62.00  & 61.60  & 14.40  & 78.00 & 68.20  & 8.20  & 53.80  & 0.00  & 1.20  & 84.80  \\
    NIAH\_MV & 83.55  & 90.45  & 86.60  & 82.10  & 95.45 & 96.40  & 50.25 & 95.10  & 0.05  & 10.80 & 96.15  \\
    NIAH\_MQ & 95.45  & 96.15  & 96.00  & 90.70  & 96.95 & 97.00  & 62.00 & 96.20  & 0.00  & 5.35  & 98.25  \\
    VT       & 57.72  & 58.56  & 56.48  & 8.92   & 96.64 & 98.16  & 25.68 & 85.72  & 0.00  & 2.92  & 97.00  \\
    CWE      & 78.20  & 75.90  & 78.20  & 73.56  & 81.38 & 80.86  & 58.78 & 82.60  & 64.70 & 23.16 & 74.26  \\
    FWE      & 84.33  & 84.20  & 84.93  & 80.07  & 58.40 & 85.53  & 26.20 & 52.60  & 18.53 & 84.93 & 81.40  \\
    QA\_1    & 62.20  & 60.40  & 62.40  & 60.60  & 57.80 & 62.40  & 60.20 & 55.80  & 26.20 & 37.20 & 55.80  \\
    QA\_2    & 43.00  & 43.40  & 42.40  & 40.20  & 42.40 & 46.20  & 43.20 & 38.20  & 28.00 & 28.20 & 46.00  \\
    \midrule
    Avg.     & 80.94  & 81.05  & 81.14  & 65.03  & 84.56 & 86.58  & 50.18 & 79.12  & 10.58 & 22.37 & 86.60 \\
    \bottomrule
    \end{tabular}
    } 
    \label{tab:ruler_4k}
\end{table*}

\begin{table*}[h]
    \centering
    \caption{Ruler results on 8k context length.}
    \resizebox{\textwidth}{!}{ 
    \begin{tabular}{l|rrrr|rrrrrrr}
    \toprule
    Exact& \multicolumn{4}{c|}{Frozen} & \multicolumn{7}{c}{Finetuned} \\
         & Base & Inf & N-F & SE & PI & N-32 & YaRN & CLEX & LLR & Land & N-64 \\
         & \checkmark & & \checkmark & & \checkmark & \checkmark & \checkmark & \checkmark & & & \checkmark \\
    \midrule
    Train Len & 4k & 4k & 4k & 4k & 32k & 32k & 32k & 32k & 32k & 32k & 64k \\
    Eval Len  & 4k & 4k & 4k & 4k & 4k  & 4k  & 4k  & 4k & 4k  & 4k  & 4k  \\
    \midrule
    NIAH\_S1 & - & 46.00  & 61.60  & 100.00 & 99.00  & 99.80  & 100.00 & 100.00 & 0.00   & 46.00  & 100.00 \\
    NIAH\_S2 & - & 36.60  & 59.40  & 98.80  & 100.00 & 100.00 & 99.40  & 100.00 & 0.00   & 7.20   & 100.00 \\
    NIAH\_S3 & - & 20.80  & 51.00  & 88.60  & 99.20  & 94.20  & 96.00  & 97.80  & 0.00   & 3.80   & 99.20  \\
    NIAH\_M1 & - & 27.80  & 46.00  & 69.40  & 98.00  & 94.20  & 86.60  & 90.20  & 0.00   & 7.60   & 95.20  \\
    NIAH\_M2 & - & 4.40   & 11.00  & 8.20   & 91.60  & 86.20  & 60.60  & 66.00  & 0.00   & 1.60   & 86.60  \\
    NIAH\_M3 & - & 2.60   & 4.00   & 3.20   & 48.40  & 52.20  & 34.60  & 11.80  & 0.00   & 0.00   & 47.40  \\
    NIAH\_MV & - & 30.35  & 41.35  & 52.95  & 65.50  & 85.95  & 70.40  & 61.25  & 0.00   & 6.25   & 84.75  \\
    NIAH\_MQ & - & 30.15  & 50.40  & 78.70  & 93.25  & 95.20  & 92.45  & 86.95  & 0.00   & 3.35   & 94.95  \\
    VT       & - & 4.88   & 69.88  & 1.48   & 91.20  & 96.16  & 77.52  & 48.16  & 0.00   & 3.08   & 94.36  \\
    CWE      & - & 65.08  & 40.30  & 30.82  & 45.66  & 45.76  & 44.72  & 32.72  & 18.92  & 22.08  & 40.80  \\
    FWE      & - & 56.73  & 64.87  & 59.00  & 65.07  & 70.13  & 10.53  & 45.40  & 16.73  & 76.60  & 54.13  \\
    QA\_1    & - & 35.80  & 44.40  & 31.00  & 50.80  & 49.20  & 43.20  & 48.20  & 22.80  & 25.00  & 50.20  \\
    QA\_2    & - & 29.00  & 33.60  & 37.40  & 40.80  & 41.80  & 36.80  & 42.60  & 24.40  & 25.20  & 44.80  \\
    \midrule
    Avg.     & - & 30.01  & 44.45  & 50.73  & 76.04  & 77.75  & 65.60  & 63.93  & 6.37   & 17.52  & 76.34  \\
    \bottomrule
    \end{tabular}
    } 
    \label{tab:ruler_8k}
\end{table*}

\begin{table*}[h]
    \centering
    \caption{Ruler results on 16k context length.}
    \resizebox{\textwidth}{!}{ 
    \begin{tabular}{l|rrrr|rrrrrrr}
    \toprule
    Exact& \multicolumn{4}{c|}{Frozen} & \multicolumn{7}{c}{Finetuned} \\
         & Base & Inf & N-F & SE & PI & N-32 & YaRN & CLEX & LLR & Land & N-64 \\
         & \checkmark & & \checkmark & & \checkmark & \checkmark & \checkmark & \checkmark & & & \checkmark \\
    \midrule
    Train Len & 4k & 4k & 4k & 4k & 32k & 32k & 32k & 32k & 32k & 32k & 64k \\
    Eval Len  & 4k & 4k & 4k & 4k & 4k  & 4k  & 4k  & 4k & 4k  & 4k  & 4k  \\
    \midrule
    NIAH\_S1 & - & 21.00  & 14.20  & 99.80  & 97.20  & 99.40  & 100.00 & 99.80  & 0.00   & 42.40  & 99.80  \\
    NIAH\_S2 & - & 17.00  & 17.40  & 93.40  & 100.00 & 100.00 & 99.20  & 100.00 & 0.20   & 6.80   & 100.00 \\
    NIAH\_S3 & - & 11.60  & 8.20   & 77.00  & 99.60  & 98.60  & 89.60  & 99.60  & 0.00   & 3.60   & 100.00 \\
    NIAH\_M1 & - & 15.80  & 9.20   & 60.00  & 97.80  & 93.20  & 83.40  & 89.40  & 0.00   & 5.60   & 90.80  \\
    NIAH\_M2 & - & 0.00   & 0.60   & 3.80   & 82.80  & 79.80  & 19.60  & 72.00  & 0.00   & 0.80   & 67.60  \\
    NIAH\_M3 & - & 1.00   & 0.00   & 1.80   & 34.20  & 18.20  & 7.40   & 15.00  & 0.00   & 0.00   & 29.60  \\
    NIAH\_MV & - & 8.40   & 6.90   & 38.85  & 77.55  & 81.95  & 58.75  & 62.40  & 0.00   & 4.80   & 83.50  \\
    NIAH\_MQ & - & 8.85   & 7.95   & 59.30  & 90.95  & 86.20  & 85.15  & 81.60  & 0.00   & 2.75   & 90.35  \\
    VT       & - & 6.56   & 11.28  & 1.16   & 68.84  & 83.56  & 47.12  & 48.16  & 0.00   & 2.52   & 88.68  \\
    CWE      & - & 19.94  & 28.36  & 17.80  & 27.26  & 26.32  & 23.72  & 28.60  & 0.62   & 11.90  & 21.20  \\
    FWE      & - & 77.13  & 25.80  & 59.80  & 47.93  & 61.73  & 10.13  & 57.33  & 12.93  & 81.60  & 51.73  \\
    QA\_1    & - & 22.80  & 36.40  & 28.00  & 46.00  & 45.20  & 43.20  & 49.20  & 13.20  & 23.00  & 45.00  \\
    QA\_2    & - & 24.20  & 26.00  & 31.60  & 35.20  & 36.00  & 37.40  & 33.40  & 20.80  & 26.20  & 36.00  \\
    \midrule
    Avg.     & - & 18.02  & 14.79  & 44.02  & 69.64  & 70.01  & 54.21  & 64.35  & 3.67   & 16.31  & 69.56  \\
    \bottomrule
    \end{tabular}
    } 
    \label{tab:ruler_16k}
\end{table*}

\begin{table*}[h]
    \centering
    \caption{Ruler results on 32k context length.}
    \resizebox{\textwidth}{!}{ 
    \begin{tabular}{l|rrrr|rrrrrrr}
    \toprule
    Exact& \multicolumn{4}{c|}{Frozen} & \multicolumn{7}{c}{Finetuned} \\
         & Base & Inf & N-F & SE & PI & N-32 & YaRN & CLEX & LLR & Land & N-64 \\
         & \checkmark & & \checkmark & & \checkmark & \checkmark & \checkmark & \checkmark & & & \checkmark \\
    \midrule
    Train Len & 4k & 4k & 4k & 4k & 32k & 32k & 32k & 32k & 32k & 32k & 64k \\
    Eval Len  & 4k & 4k & 4k & 4k & 4k  & 4k  & 4k  & 4k & 4k  & 4k  & 4k  \\
    \midrule
    NIAH\_S1 & - & 7.80  & 0.00 & 83.00 & 97.20 & 99.00  & 85.80 & 85.60 & 0.00  & 33.80 & 100.00 \\
    NIAH\_S2 & - & 7.00  & 0.00 & 68.40 & 99.00 & 100.00 & 81.00 & 94.00 & 0.00  & 3.20  & 99.20  \\
    NIAH\_S3 & - & 6.40  & 0.00 & 42.80 & 97.00 & 99.40  & 62.40 & 97.20 & 0.00  & 2.60  & 96.40  \\
    NIAH\_M1 & - & 8.80  & 0.00 & 29.40 & 93.40 & 90.80  & 63.20 & 78.40 & 0.00  & 5.40  & 82.60  \\
    NIAH\_M2 & - & 0.00  & 0.00 & 2.40  & 48.80 & 39.40  & 6.40  & 40.40 & 0.00  & 0.20  & 36.60  \\
    NIAH\_M3 & - & 0.00  & 0.00 & 1.40  & 5.80  & 8.60   & 1.20  & 8.00  & 0.00  & 0.00  & 7.20   \\
    NIAH\_MV & - & 3.75  & 0.00 & 24.65 & 61.30 & 68.20  & 37.95 & 60.75 & 0.00  & 2.80  & 82.20  \\
    NIAH\_MQ & - & 2.05  & 0.00 & 20.35 & 68.65 & 78.25  & 46.95 & 67.80 & 0.05  & 2.35  & 85.80  \\
    VT       & - & 2.08  & 0.00 & 2.32  & 56.68 & 43.28  & 22.00 & 30.08 & 0.00  & 2.52  & 71.28  \\
    CWE      & - & 4.48  & 0.02 & 17.46 & 26.72 & 11.78  & 11.38 & 22.70 & 13.26 & 3.68  & 7.34   \\
    FWE      & - & 72.67 & 2.93 & 46.73 & 31.00 & 64.53  & 15.13 & 34.67 & 13.93 & 72.47 & 51.53  \\
    QA\_1    & - & 20.20 & 5.20 & 20.60 & 33.40 & 34.40  & 23.00 & 28.00 & 6.00  & 22.60 & 27.00  \\
    QA\_2    & - & 25.20 & 1.20 & 24.00 & 30.60 & 34.80  & 24.00 & 30.60 & 12.60 & 24.60 & 33.20  \\
    \midrule
    Avg.     & - & 12.34 & 0.72 & 29.50 & 57.66 & 59.42  & 36.95 & 52.17 & 3.53  & 13.56 & 60.03 \\
    \bottomrule
    \end{tabular}
    } 
    \label{tab:ruler_32k}
\end{table*}

\begin{table*}[h]
    \centering
    \caption{Ruler results on 64k context length.}
    \resizebox{\textwidth}{!}{ 
    \begin{tabular}{l|rrrr|rrrrrrr}
    \toprule
    Exact& \multicolumn{4}{c|}{Frozen} & \multicolumn{7}{c}{Finetuned} \\
         & Base & Inf & N-F & SE & PI & N-32 & YaRN & CLEX & LLR & Land & N-64 \\
         & \checkmark & & \checkmark & & \checkmark & \checkmark & \checkmark & \checkmark & & & \checkmark \\
    \midrule
    Train Len & 4k & 4k & 4k & 4k & 32k & 32k & 32k & 32k & 32k & 32k & 64k \\
    Eval Len  & 4k & 4k & 4k & 4k & 4k  & 4k  & 4k  & 4k & 4k  & 4k  & 4k  \\
    \midrule
    NIAH\_S1 & - & 3.20  & 0.00  & 71.80 & 0.00 & 83.60 & 0.00 & 40.60 & 0.00 & 40.00 & 98.00 \\
    NIAH\_S2 & - & 3.80  & 0.00  & 0.20  & 0.00 & 95.60 & 0.00 & 68.80 & 0.00 & 3.00  & 98.00 \\
    NIAH\_S3 & - & 5.40  & 0.00  & 0.00  & 0.00 & 95.40 & 0.00 & 70.40 & 0.00 & 3.00  & 95.80 \\
    NIAH\_M1 & - & 5.40  & 0.00  & 0.00  & 0.00 & 76.80 & 0.00 & 55.40 & 0.00 & 5.20  & 67.20 \\
    NIAH\_M2 & - & 0.00  & 0.00  & 2.60  & 0.00 & 15.20 & 0.00 & 15.80 & 0.00 & 0.00  & 25.80 \\
    NIAH\_M3 & - & 0.00  & 0.00  & 0.20  & 0.00 & 1.20  & 0.00 & 1.00  & 0.00 & 0.00  & 4.00  \\
    NIAH\_MV & - & 4.45  & 0.00  & 0.20  & 0.00 & 51.70 & 0.00 & 36.40 & 0.00 & 3.70  & 51.20 \\
    NIAH\_MQ & - & 4.45  & 0.00  & 0.05  & 0.00 & 56.60 & 0.00 & 43.50 & 0.00 & 2.45  & 65.40 \\
    VT       & - & 1.28  & 0.00  & 12.20 & 0.00 & 34.28 & 0.00 & 0.00  & 0.00 & 2.40  & 41.48 \\
    CWE      & - & 0.76  & 0.00  & 6.85  & 0.00 & 6.58  & 0.00 & 9.72  & 0.00 & 1.70  & 7.88  \\
    FWE      & - & 72.20 & 11.47 & 26.47 & 0.00 & 25.27 & 0.00 & 11.73 & 0.00 & 82.67 & 27.73 \\
    QA\_1    & - & 16.20 & 0.20  & 0.80  & 0.00 & 30.80 & 0.00 & 25.60 & 0.00 & 19.60 & 29.20 \\
    QA\_2    & - & 20.20 & 0.20  & 0.00  & 0.00 & 28.40 & 0.00 & 19.00 & 0.00 & 20.20 & 29.40 \\
    \midrule
    Avg.     & - & 10.56 & 0.91  & 9.34  & 0.00 & 46.26 & 0.00 & 30.61 & 0.00 & 14.15 & 49.31 \\
    \bottomrule
    \end{tabular}
    } 
    \label{tab:ruler_64k}
\end{table*}

\clearpage

%% file: main.bbl
\begin{thebibliography}{44}
\providecommand{\natexlab}[1]{#1}
\providecommand{\url}[1]{\texttt{#1}}
\expandafter\ifx\csname urlstyle\endcsname\relax
  \providecommand{\doi}[1]{doi: #1}\else
  \providecommand{\doi}{doi: \begingroup \urlstyle{rm}\Url}\fi

\bibitem[AI@Meta(2024)]{llama3modelcard}
AI@Meta.
\newblock Llama 3 model card.
\newblock 2024.
\newblock URL \url{https://github.com/meta-llama/llama3/blob/main/MODEL_CARD.md}.

\bibitem[Liu et~al.(2023{\natexlab{a}})Liu, Zaharia, and Abbeel]{liu2023ring}
Hao Liu, Matei Zaharia, and Pieter Abbeel.
\newblock Ring attention with blockwise transformers for near-infinite context.
\newblock In \emph{NeurIPS 2023 Foundation Models for Decision Making Workshop}, 2023{\natexlab{a}}.
\newblock URL \url{https://openreview.net/forum?id=fXugVDtCQO}.

\bibitem[Tanzer et~al.(2024)Tanzer, Suzgun, Visser, Jurafsky, and Melas-Kyriazi]{tanzer2024a}
Garrett Tanzer, Mirac Suzgun, Eline Visser, Dan Jurafsky, and Luke Melas-Kyriazi.
\newblock A benchmark for learning to translate a new language from one grammar book.
\newblock In \emph{The Twelfth International Conference on Learning Representations}, 2024.
\newblock URL \url{https://openreview.net/forum?id=tbVWug9f2h}.

\bibitem[Kry{\'s}ci{\'n}ski et~al.(2022)Kry{\'s}ci{\'n}ski, Rajani, Agarwal, Xiong, and Radev]{kryscinski2022booksum}
Wojciech Kry{\'s}ci{\'n}ski, Nazneen Rajani, Divyansh Agarwal, Caiming Xiong, and Dragomir Radev.
\newblock Booksum: A collection of datasets for long-form narrative summarization.
\newblock In \emph{Findings of the Association for Computational Linguistics: EMNLP 2022}, pages 6536--6558, 2022.

\bibitem[Bertsch et~al.(2024)Bertsch, Ivgi, Alon, Berant, Gormley, and Neubig]{bertsch2024context}
Amanda Bertsch, Maor Ivgi, Uri Alon, Jonathan Berant, Matthew~R Gormley, and Graham Neubig.
\newblock In-context learning with long-context models: An in-depth exploration.
\newblock \emph{arXiv preprint arXiv:2405.00200}, 2024.

\bibitem[Li et~al.(2023{\natexlab{a}})Li, Gong, Feng, Xu, Zhang, Wu, and Kong]{li2023context}
Mukai Li, Shansan Gong, Jiangtao Feng, Yiheng Xu, Jun Zhang, Zhiyong Wu, and Lingpeng Kong.
\newblock In-context learning with many demonstration examples.
\newblock \emph{arXiv preprint arXiv:2302.04931}, 2023{\natexlab{a}}.

\bibitem[Chen et~al.(2023{\natexlab{a}})Chen, Wong, Chen, and Tian]{chen2023extending}
Shouyuan Chen, Sherman Wong, Liangjian Chen, and Yuandong Tian.
\newblock Extending context window of large language models via positional interpolation, 2023{\natexlab{a}}.

\bibitem[Peng et~al.(2023)Peng, Quesnelle, Fan, and Shippole]{peng2023yarn}
Bowen Peng, Jeffrey Quesnelle, Honglu Fan, and Enrico Shippole.
\newblock Yarn: Efficient context window extension of large language models, 2023.

\bibitem[Han et~al.(2023)Han, Wang, Xiong, Chen, Ji, and Wang]{han2023lminfinite}
Chi Han, Qifan Wang, Wenhan Xiong, Yu~Chen, Heng Ji, and Sinong Wang.
\newblock Lm-infinite: Simple on-the-fly length generalization for large language models, 2023.

\bibitem[bloc97(2023)]{bloc97}
bloc97.
\newblock {NTK-Aware Scaled RoPE allows LLaMA models to have extended (8k+) context size without any fine-tuning and minimal perplexity degradation}, 2023.
\newblock URL \url{https://www.reddit.com/r/LocalLLaMA/comments/14lz7j5/ntkaware_scaled_rope_allows_llama_models_to_have/}.

\bibitem[Chen et~al.(2023{\natexlab{b}})Chen, Qian, Tang, Lai, Liu, Han, and Jia]{chen2023longlora}
Yukang Chen, Shengju Qian, Haotian Tang, Xin Lai, Zhijian Liu, Song Han, and Jiaya Jia.
\newblock Longlora: Efficient fine-tuning of long-context large language models, 2023{\natexlab{b}}.

\bibitem[Mohtashami and Jaggi(2023)]{mohtashami2023landmark}
Amirkeivan Mohtashami and Martin Jaggi.
\newblock Landmark attention: Random-access infinite context length for transformers, 2023.

\bibitem[gkamradt(2023)]{NeedleInAHaystack}
gkamradt.
\newblock Needle in a haystack - pressure testing llms, 2023.
\newblock URL \url{https://github.com/gkamradt/LLMTest_NeedleInAHaystack}.

\bibitem[Bai et~al.(2023)Bai, Lv, Zhang, Lyu, Tang, Huang, Du, Liu, Zeng, Hou, Dong, Tang, and Li]{bai2023longbench}
Yushi Bai, Xin Lv, Jiajie Zhang, Hongchang Lyu, Jiankai Tang, Zhidian Huang, Zhengxiao Du, Xiao Liu, Aohan Zeng, Lei Hou, Yuxiao Dong, Jie Tang, and Juanzi Li.
\newblock Longbench: A bilingual, multitask benchmark for long context understanding, 2023.

\bibitem[An et~al.(2023)An, Gong, Zhong, Zhao, Li, Zhang, Kong, and Qiu]{an2023leval}
Chenxin An, Shansan Gong, Ming Zhong, Xingjian Zhao, Mukai Li, Jun Zhang, Lingpeng Kong, and Xipeng Qiu.
\newblock L-eval: Instituting standardized evaluation for long context language models, 2023.

\bibitem[Xiao et~al.(2024)Xiao, Zhang, Han, Xiao, Lin, Zhang, Liu, Han, and Sun]{xiao2024infllm}
Chaojun Xiao, Pengle Zhang, Xu~Han, Guangxuan Xiao, Yankai Lin, Zhengyan Zhang, Zhiyuan Liu, Song Han, and Maosong Sun.
\newblock Infllm: Unveiling the intrinsic capacity of llms for understanding extremely long sequences with training-free memory.
\newblock \emph{arXiv preprint arXiv:2402.04617}, 2024.

\bibitem[Lu et~al.(2024)Lu, Zhou, He, Zhao, Ji, Gui, Zhang, and Huang]{lu2024longheads}
Yi~Lu, Xin Zhou, Wei He, Jun Zhao, Tao Ji, Tao Gui, Qi~Zhang, and Xuanjing Huang.
\newblock Longheads: Multi-head attention is secretly a long context processor.
\newblock \emph{arXiv preprint arXiv:2402.10685}, 2024.

\bibitem[Fu et~al.(2024)Fu, Panda, Niu, Yue, Hajishirzi, Kim, and Peng]{fu2024data}
Yao Fu, Rameswar Panda, Xinyao Niu, Xiang Yue, Hannaneh Hajishirzi, Yoon Kim, and Hao Peng.
\newblock Data engineering for scaling language models to 128k context, 2024.

\bibitem[Hsieh et~al.(2024)Hsieh, Sun, Kriman, Acharya, Rekesh, Jia, Zhang, and Ginsburg]{hsieh2024ruler}
Cheng-Ping Hsieh, Simeng Sun, Samuel Kriman, Shantanu Acharya, Dima Rekesh, Fei Jia, Yang Zhang, and Boris Ginsburg.
\newblock Ruler: What's the real context size of your long-context language models?, 2024.

\bibitem[emozilla(2023)]{emozillareddit}
emozilla.
\newblock {Dynamically Scaled RoPE further increases performance of long context LLaMA with zero fine-tuning}, 2023.
\newblock URL \url{https://www.reddit.com/r/LocalLLaMA/comments/14mrgpr/dynamically_scaled_rope_further_increases/}.

\bibitem[Chen et~al.(2024)Chen, Li, Meng, Liang, and Bing]{chen2024clexcontinuouslengthextrapolation}
Guanzheng Chen, Xin Li, Zaiqiao Meng, Shangsong Liang, and Lidong Bing.
\newblock Clex: Continuous length extrapolation for large language models, 2024.
\newblock URL \url{https://arxiv.org/abs/2310.16450}.

\bibitem[Su et~al.(2021)Su, Lu, Pan, Wen, and Liu]{Su_Lu_Pan_Wen_Liu_2021}
Jianlin Su, Yu~Lu, Shengfeng Pan, Bo~Wen, and Yunfeng Liu.
\newblock Roformer: Enhanced transformer with rotary position embedding.
\newblock \emph{Cornell University - arXiv,Cornell University - arXiv}, Apr 2021.

\bibitem[Tworkowski et~al.(2023)Tworkowski, Staniszewski, Pacek, Wu, Michalewski, and Miłoś]{tworkowski2023focused}
Szymon Tworkowski, Konrad Staniszewski, Mikołaj Pacek, Yuhuai Wu, Henryk Michalewski, and Piotr Miłoś.
\newblock Focused transformer: Contrastive training for context scaling, 2023.

\bibitem[Hu et~al.(2021)Hu, Shen, Wallis, Allen{-}Zhu, Li, Wang, and Chen]{DBLP:journals/corr/abs-2106-09685}
Edward~J. Hu, Yelong Shen, Phillip Wallis, Zeyuan Allen{-}Zhu, Yuanzhi Li, Shean Wang, and Weizhu Chen.
\newblock Lora: Low-rank adaptation of large language models.
\newblock \emph{CoRR}, abs/2106.09685, 2021.
\newblock URL \url{https://arxiv.org/abs/2106.09685}.

\bibitem[Xiao et~al.(2023)Xiao, Tian, Chen, Han, and Lewis]{xiao2023efficient}
Guangxuan Xiao, Yuandong Tian, Beidi Chen, Song Han, and Mike Lewis.
\newblock Efficient streaming language models with attention sinks, 2023.

\bibitem[Xu et~al.(2024)Xu, Ping, Wu, McAfee, Zhu, Liu, Subramanian, Bakhturina, Shoeybi, and Catanzaro]{xu2024retrieval}
Peng Xu, Wei Ping, Xianchao Wu, Lawrence McAfee, Chen Zhu, Zihan Liu, Sandeep Subramanian, Evelina Bakhturina, Mohammad Shoeybi, and Bryan Catanzaro.
\newblock Retrieval meets long context large language models, 2024.

\bibitem[Jiang et~al.(2023)Jiang, Wu, Luo, Li, Lin, Yang, and Qiu]{jiang2023longllmlingua}
Huiqiang Jiang, Qianhui Wu, Xufang Luo, Dongsheng Li, Chin-Yew Lin, Yuqing Yang, and Lili Qiu.
\newblock Longllmlingua: Accelerating and enhancing llms in long context scenarios via prompt compression, 2023.

\bibitem[Li et~al.(2023{\natexlab{b}})Li, Dong, Lin, and Guerin]{li2023compressing}
Yucheng Li, Bo~Dong, Chenghua Lin, and Frank Guerin.
\newblock Compressing context to enhance inference efficiency of large language models, 2023{\natexlab{b}}.

\bibitem[Tay et~al.(2020)Tay, Dehghani, Abnar, Shen, Bahri, Pham, Rao, Yang, Ruder, and Metzler]{tay2020long}
Yi~Tay, Mostafa Dehghani, Samira Abnar, Yikang Shen, Dara Bahri, Philip Pham, Jinfeng Rao, Liu Yang, Sebastian Ruder, and Donald Metzler.
\newblock Long range arena: A benchmark for efficient transformers, 2020.

\bibitem[Li et~al.(2023{\natexlab{c}})Li, Wang, Zheng, and Zhang]{li2023loogle}
Jiaqi Li, Mengmeng Wang, Zilong Zheng, and Muhan Zhang.
\newblock Loogle: Can long-context language models understand long contexts?, 2023{\natexlab{c}}.

\bibitem[Liu et~al.(2023{\natexlab{b}})Liu, Lin, Hewitt, Paranjape, Bevilacqua, Petroni, and Liang]{liu2023lost}
Nelson~F. Liu, Kevin Lin, John Hewitt, Ashwin Paranjape, Michele Bevilacqua, Fabio Petroni, and Percy Liang.
\newblock Lost in the middle: How language models use long contexts, 2023{\natexlab{b}}.

\bibitem[Su et~al.(2024)Su, Ahmed, Lu, Pan, Bo, and Liu]{su2024roformer}
Jianlin Su, Murtadha Ahmed, Yu~Lu, Shengfeng Pan, Wen Bo, and Yunfeng Liu.
\newblock Roformer: Enhanced transformer with rotary position embedding.
\newblock \emph{Neurocomputing}, 568:\penalty0 127063, 2024.

\bibitem[Together.AI(2023)]{together-instruct}
Together.AI.
\newblock Llama-2-7b-32k-instruct — and fine-tuning for llama-2 models with together api, 2023.
\newblock URL \url{https://www.together.ai/blog/llama-2-7b-32k-instruct}.

\bibitem[Chiang et~al.(2023)Chiang, Li, Lin, Sheng, Wu, Zhang, Zheng, Zhuang, Zhuang, Gonzalez, Stoica, and Xing]{vicuna2023}
Wei-Lin Chiang, Zhuohan Li, Zi~Lin, Ying Sheng, Zhanghao Wu, Hao Zhang, Lianmin Zheng, Siyuan Zhuang, Yonghao Zhuang, Joseph~E. Gonzalez, Ion Stoica, and Eric~P. Xing.
\newblock Vicuna: An open-source chatbot impressing gpt-4 with 90\%* chatgpt quality, March 2023.
\newblock URL \url{https://lmsys.org/blog/2023-03-30-vicuna/}.

\bibitem[Jin et~al.(2024)Jin, Han, Yang, Jiang, Liu, Chang, Chen, and Hu]{jin2024llm}
Hongye Jin, Xiaotian Han, Jingfeng Yang, Zhimeng Jiang, Zirui Liu, Chia-Yuan Chang, Huiyuan Chen, and Xia Hu.
\newblock Llm maybe longlm: Self-extend llm context window without tuning, 2024.

\bibitem[Touvron et~al.(2023)Touvron, Martin, Stone, Albert, Almahairi, Babaei, Bashlykov, Batra, Bhargava, Bhosale, Bikel, Blecher, Ferrer, Chen, Cucurull, Esiobu, Fernandes, Fu, Fu, Fuller, Gao, Goswami, Goyal, Hartshorn, Hosseini, Hou, Inan, Kardas, Kerkez, Khabsa, Kloumann, Korenev, Koura, Lachaux, Lavril, Lee, Liskovich, Lu, Mao, Martinet, Mihaylov, Mishra, Molybog, Nie, Poulton, Reizenstein, Rungta, Saladi, Schelten, Silva, Smith, Subramanian, Tan, Tang, Taylor, Williams, Kuan, Xu, Yan, Zarov, Zhang, Fan, Kambadur, Narang, Rodriguez, Stojnic, Edunov, and Scialom]{touvron2023llama}
Hugo Touvron, Louis Martin, Kevin Stone, Peter Albert, Amjad Almahairi, Yasmine Babaei, Nikolay Bashlykov, Soumya Batra, Prajjwal Bhargava, Shruti Bhosale, Dan Bikel, Lukas Blecher, Cristian~Canton Ferrer, Moya Chen, Guillem Cucurull, David Esiobu, Jude Fernandes, Jeremy Fu, Wenyin Fu, Brian Fuller, Cynthia Gao, Vedanuj Goswami, Naman Goyal, Anthony Hartshorn, Saghar Hosseini, Rui Hou, Hakan Inan, Marcin Kardas, Viktor Kerkez, Madian Khabsa, Isabel Kloumann, Artem Korenev, Punit~Singh Koura, Marie-Anne Lachaux, Thibaut Lavril, Jenya Lee, Diana Liskovich, Yinghai Lu, Yuning Mao, Xavier Martinet, Todor Mihaylov, Pushkar Mishra, Igor Molybog, Yixin Nie, Andrew Poulton, Jeremy Reizenstein, Rashi Rungta, Kalyan Saladi, Alan Schelten, Ruan Silva, Eric~Michael Smith, Ranjan Subramanian, Xiaoqing~Ellen Tan, Binh Tang, Ross Taylor, Adina Williams, Jian~Xiang Kuan, Puxin Xu, Zheng Yan, Iliyan Zarov, Yuchen Zhang, Angela Fan, Melanie Kambadur, Sharan Narang, Aurelien Rodriguez, Robert Stojnic, Sergey Edunov, and Thomas
  Scialom.
\newblock Llama 2: Open foundation and fine-tuned chat models, 2023.

\bibitem[Javaheripi et~al.(2023)Javaheripi, Bubeck, Abdin, Aneja, Bubeck, Mendes, Chen, Giorno, Eldan, Gopi, Gunasekar, Javaheripi, Kauffmann, Lee, Li, Nguyen, de~Rosa, Saarikivi, Salim, Shah, Santacroce, Behl, Kalai, Wang, Ward, Witte, Zhang, and Zhang]{javaheripi2023phi}
Mojan Javaheripi, Sébastien Bubeck, Marah Abdin, Jyoti Aneja, Sebastien Bubeck, Caio César~Teodoro Mendes, Weizhu Chen, Allie~Del Giorno, Ronen Eldan, Sivakanth Gopi, Suriya Gunasekar, Mojan Javaheripi, Piero Kauffmann, Yin~Tat Lee, Yuanzhi Li, Anh Nguyen, Gustavo de~Rosa, Olli Saarikivi, Adil Salim, Shital Shah, Michael Santacroce, Harkirat~Singh Behl, Adam~Taumann Kalai, Xin Wang, Rachel Ward, Philipp Witte, Cyril Zhang, and Yi~Zhang.
\newblock Phi-2: The surprising power of small language models, 2023.

\bibitem[Rae et~al.(2019)Rae, Potapenko, Jayakumar, and Lillicrap]{rae2019compressive}
Jack~W. Rae, Anna Potapenko, Siddhant~M. Jayakumar, and Timothy~P. Lillicrap.
\newblock Compressive transformers for long-range sequence modelling, 2019.

\bibitem[Azerbayev et~al.(2023)Azerbayev, Piotrowski, Schoelkopf, Ayers, Radev, and Avigad]{azerbayev2023proofnet}
Zhangir Azerbayev, Bartosz Piotrowski, Hailey Schoelkopf, Edward~W. Ayers, Dragomir Radev, and Jeremy Avigad.
\newblock Proofnet: Autoformalizing and formally proving undergraduate-level mathematics, 2023.

\bibitem[Press et~al.(2022)Press, Smith, and Lewis]{press2022train}
Ofir Press, Noah Smith, and Mike Lewis.
\newblock Train short, test long: Attention with linear biases enables input length extrapolation.
\newblock In \emph{International Conference on Learning Representations}, 2022.
\newblock URL \url{https://openreview.net/forum?id=R8sQPpGCv0}.

\bibitem[Kontonis et~al.(2024)Kontonis, Ma, and Tzamos]{kontonis2024active}
Vasilis Kontonis, Mingchen Ma, and Christos Tzamos.
\newblock Active learning with simple questions, 2024.

\bibitem[Li and Roth(2002)]{Li_Roth_2002}
Xin Li and Dan Roth.
\newblock Learning question classifiers.
\newblock In \emph{Proceedings of the 19th international conference on Computational linguistics -}, Jan 2002.
\newblock \doi{10.3115/1072228.1072378}.
\newblock URL \url{http://dx.doi.org/10.3115/1072228.1072378}.

\bibitem[Sun et~al.(2021)Sun, Krishna, Mattarella-Micke, and Iyyer]{sun2021longrange}
Simeng Sun, Kalpesh Krishna, Andrew Mattarella-Micke, and Mohit Iyyer.
\newblock Do long-range language models actually use long-range context?, 2021.

\bibitem[Soboleva et~al.(2023)Soboleva, Al-Khateeb, Myers, Steeves, Hestness, and Dey]{cerebras2023slimpajama}
Daria Soboleva, Faisal Al-Khateeb, Robert Myers, Jacob~R Steeves, Joel Hestness, and Nolan Dey.
\newblock {SlimPajama: A 627B token cleaned and deduplicated version of RedPajama}.
\newblock \url{https://www.cerebras.net/blog/slimpajama-a-627b-token-cleaned-and-deduplicated-version-of-redpajama}, 2023.
\newblock URL \url{https://huggingface.co/datasets/cerebras/SlimPajama-627B}.

\end{thebibliography}
